\documentclass[10pt,twocolumn,letterpaper]{article}

\usepackage{cvpr}              
\usepackage[accsupp]{axessibility}  
\usepackage{epigraph}
\definecolor{cvprblue}{rgb}{0.21,0.49,0.74}
\usepackage[pagebackref,breaklinks,colorlinks,allcolors=cvprblue]{hyperref}

\def\paperID{****} 
\def\confName{CVPR}
\def\confYear{2026}

\usepackage{caption}
\usepackage{graphicx}
\usepackage{multirow}    
\usepackage{booktabs}    
\usepackage{amssymb}
\usepackage{pifont}
\usepackage[table]{xcolor}
\usepackage{tikz}
\usepackage{arydshln}

\usepackage{stfloats}
\usepackage{cuted}

\newcommand{\cmark}{\textcolor{green}{\checkmark}}

\usepackage{tcolorbox}
\tcbuselibrary{skins}
\newcounter{prompt}[section] 
\renewcommand{\theprompt}{\thesection.\arabic{prompt}}
\newtcolorbox{promptbox}[2][]{%
  promptstyle,
  before upper={%
    \refstepcounter{prompt}
    \noindent\textbf{\color{white}\fcolorbox{black}{black}{Prompt~\theprompt: #2}}%
  },
  #1
}
\tcbset{
  promptstyle/.style={
    colback=gray!6, colframe=black,
    left=2mm, right=2mm, top=1mm, bottom=1mm,
    sharp corners, boxrule=0.4pt,
    fonttitle=\bfseries,
    colbacktitle=black,       
    coltitle=white             
  }
}

\title{RealVLG-R1: A Large-Scale Real-World Visual-Language Grounding Benchmark for Robotic Perception and Manipulation}

\begin{document}

\author{
Linfei Li \quad
Lin Zhang\footnotemark[1] \quad
Ying Shen \\
School of Computer Science and Technology, Tongji University, China \\
{\tt\small \{cslinfeili, cslinzhang, yingshen\}@tongji.edu.cn}
}

\twocolumn[{
\maketitle
\begin{center}
    \vspace{-1.5em} 
    \includegraphics[width=0.94\linewidth]{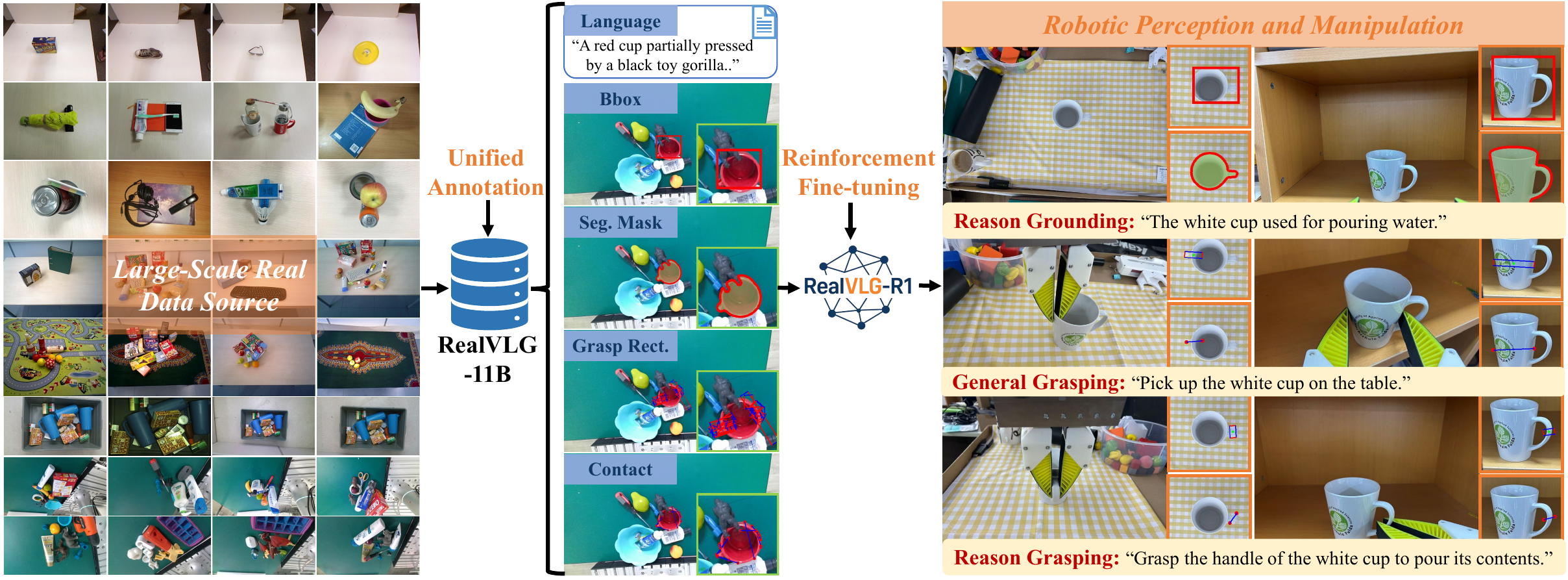}
    \vspace{-0.8em}
    \captionsetup{type=figure}
    \captionof{figure}{
        We propose RealVLG, a unified framework that integrates the RealVLG-11B dataset and RealVLG-R1 model to enable multi-granularity, zero-shot robotic visual-language grounding and grasping in real-world scenarios.
    }
    \label{fig:teaser}
    \vspace{-8px}
\end{center}
}]

{
\renewcommand{\thefootnote}{\fnsymbol{footnote}}
\footnotetext[1]{Corresponding author: cslinzhang@tongji.edu.cn}
}

\begin{abstract}
Visual-language grounding aims to establish semantic correspondences between natural language and visual entities, enabling models to accurately identify and localize target objects based on textual instructions. Existing VLG approaches focus on coarse-grained, object-level localization, while traditional robotic grasping methods rely predominantly on geometric cues and lack language guidance, which limits their applicability in language-driven manipulation scenarios. To address these limitations, we propose the RealVLG framework, which integrates the RealVLG-11B dataset and the RealVLG-R1 model to unify real-world visual-language grounding and grasping tasks. RealVLG-11B dataset provides multi-granularity annotations including bounding boxes, segmentation masks, grasp poses, contact points, and human-verified fine-grained language descriptions, covering approximately 165,000 images, over 800 object instances, 1.3 million segmentation, detection, and language annotations, and roughly 11 billion grasping examples. Building on this dataset, RealVLG-R1 employs Reinforcement Fine-tuning on pretrained large-scale vision-language models to predict bounding boxes, segmentation masks, grasp poses, and contact points in a unified manner given natural language instructions. Experimental results demonstrate that RealVLG supports zero-shot perception and manipulation in real-world unseen environments, establishing a unified semantic-visual multimodal benchmark that provides a comprehensive data and evaluation platform for language-driven robotic perception and grasping policy learning. All data and code are publicly available at \url{https://github.com/lif314/RealVLG-R1}.
\end{abstract}
    
\section{Introduction}
\label{sec:introduction}
\epigraph{\textit{If I have seen further, it is by standing on the shoulders of giants.}}{Isaac Newton}

Visual-Language Grounding (VLG) aims to establish semantic correspondences between natural language and visual entities in images, enabling models to accurately identify and localize target objects based on textual instructions \cite{vlg20161, vlg20162}. As a core task in cross-modal understanding, VLG plays a fundamental role in language-driven perception, interaction, and intelligent control. Recent advances in Large Vision-Language Models (LVLMs) \cite{BLIP2023, LLaVA2024, LLAVA2023, bai2025qwen2} have significantly enhanced object-level semantic alignment, allowing models to generate bounding boxes \cite{chen2023shikra, CogVLM2025, you2023ferret, liu2025visualrft} or segmentation masks \cite{lisa2024, GLaMM2024, GSVA2024, PSALM2025,YLVLG2025} in open-world scenarios, thereby establishing an initial mapping from language to vision. This capability provides critical support for various downstream tasks, including visual question answering \cite{Chen_2024_CVPR, Yang_2025_CVPR}, staged language-guided object manipulation \cite{qian2024thinkgrasp, tziafas2024towards, meng2024ovgnet}, and task planning \cite{huang2024rekep, OmniManip2025}, further laying the foundation for robotic recognition and manipulation of actionable objects in complex environments.

However, existing research on VLG primarily focuses on coarse-grained, object-level localization and has not yet been extended to actionable grasp understanding, i.e., reasoning about how to interact with objects. Meanwhile, traditional grasping studies \cite{graspnet2020, ZeroGraspCVPR2025, anygrasp2021, Ma_2024_CVPR, grcnn2020, mahler2017dex, cornell2011, ocidgrasp2021} have achieved significant progress in geometric modeling and grasp pose estimation, but they largely rely on visual or geometric cues and lack language and semantic guidance, limiting their applicability in multi-task, human-interactive operational scenarios. In addition, existing datasets, such as Grasp-Anything \cite{graspanything2024, lgd2024}, attempt to combine visual-language information with grasp annotations; however, they consist of low-resolution, diffusion-based synthetic scenes \cite{diffusion2022}, which reduces realism. Furthermore, their grasp annotations are generated by RAGT-3/3 \cite{cao2023nbmod} and suffer from limited accuracy, while the associated language instructions are predominantly coarse-grained, covering only scene or object categories and lacking fine-grained, object-aligned descriptions or spatial relationship information. In summary, current VLG and grasping research highlight a clear gap between semantic understanding and manipulation reasoning, making them insufficient for real-world robotic scenarios that require fine-grained, multi-modal perception.

To address these challenges, as shown in Fig. \ref{fig:teaser}, we propose the RealVLG framework, which unifies visual-language grounding and grasping tasks within a single research paradigm. Building upon existing real-world grasping datasets, we construct the RealVLG-11B dataset, systematically extending traditional grasping annotations with segmentation masks, bounding boxes, contact points, and natural language descriptions. The language annotations are first automatically generated by GPT-4o \cite{openai2024gpt4o}, then validated by Qwen-VL-Max \cite{alibaba2025qwenvlmax}, and finally reviewed manually to ensure precise alignment between textual descriptions and visual objects. This dataset provides multi-granularity annotations, spanning semantic-level object localization to grasp-level operational understanding, and comprises approximately 165,000 images, over 800 object instances, roughly 1.3 million segmentation, detection, and language annotations, and around 11 billion grasping examples, making it the largest dataset to date that simultaneously captures semantic and visual information for real-world perception.

Based on this dataset, we further introduce the RealVLG-R1 model. Leveraging LVLMs (e.g., the Qwen-VL series) as the backbone, the model is trained using a Reinforcement Fine-tuning strategy to directly predict segmentation masks, bounding boxes, grasping poses, and contact points from natural language instructions. During training, input images and task prompts are processed through a policy optimization module to generate candidate outputs, which are then updated using verifiable reward signals. Finally, RealVLG-R1 demonstrates zero-shot robotic perception and manipulation in real-world scenarios, highlighting its generalization capabilities and execution performance in unseen environments.

By integrating the dataset and model, the RealVLG framework establishes a unified visual-language grounding and grasping benchmark, providing a novel data foundation and evaluation platform for language-driven robotic perception and manipulation.

In summary, our main contributions are as follows:
\begin{itemize}
    \item RealVLG-11B Dataset: The largest real-world grounding and grasping dataset with multi-granularity annotations from semantic localization to grasp-level understanding.
    \item RealVLG-R1 Model: A unified model trained via Reinforcement Learning Fine-tuning for zero-shot language-driven grounding and grasping.
    \item RealVLG Benchmark: A unified visual-language grounding and grasping benchmark for robotic perception and grasping policy learning.
\end{itemize}

\begin{table*}[ht]
\centering
\resizebox{0.90\textwidth}{!}{
\begin{tabular}{clccccccccccc}
\hline
\multirow{2}{*}{Type} & \multirow{2}{*}{Dataset} &  \multirow{2}{*}{\shortstack{Grasp \\ Label}} & \multirow{2}{*}{\shortstack{Objects \\ / Image}} & \multicolumn{3}{c}{\textbf{Scale}} &  \multicolumn{2}{c}{\textbf{Grounding}} &  \multicolumn{4}{c}{\textbf{Language Description}} \\
\cmidrule(l){5-7} \cmidrule(l){8-9} \cmidrule(l){10-13} 
 &  & & &  Images & Objects & Grasps & Seg. & Bbox & Object & Spatial & VLM-Check & Human-Check  \\
\hline
\multirow{8}{*}{Sim.} 
& Dex-net \cite{mahler2017dex} & Rect. & 1 & 6.7M & 1500 & 6.7M & \cmark & \ding{55} & \ding{55} & \ding{55} & \ding{55} & \ding{55}  \\
& Jacquard \cite{Jacquard2018} & Rect. & 1 & 54K & 11K & 1.1 M & \cmark & \ding{55} & \ding{55} & \ding{55} & \ding{55} & \ding{55} \\
& VR-Grasping-101 \cite{VRGrasping2018} & 6-DoF & 1 & 10K & 101 & 4.8M & \ding{55} & \ding{55} & \ding{55} & \ding{55} & \ding{55} & \ding{55} \\
& Kappler et al. \cite{Kappler2015} & 6-DoF & 1 & 700 & 80 & 300K & \ding{55} & \ding{55} & \ding{55} & \ding{55} & \ding{55} & \ding{55} \\
& 6-DoF GraspNet \cite{6dofgraspnet2019} & 6-DoF & 1 & 206 & 206 & 7M& \ding{55} & \ding{55} & \ding{55} & \ding{55} & \ding{55} & \ding{55} \\
& Epper et al. \cite{eppner2019billion} & 6-DoF & 1 & 21 & 21 & 1B & \ding{55} & \ding{55} & \ding{55} & \ding{55}& \ding{55} & \ding{55} \\
& EGAD! \cite{egad2020} & 6-DoF & 1 & 2231 & 2231 & 233K& \ding{55} & \ding{55} & \ding{55} & \ding{55} & \ding{55} & \ding{55} \\
& ACRONYM \cite{ACRONYM2021} & 6-DoF & - & - & 8872 & 17.7M & \ding{55} & \ding{55} & \ding{55} & \ding{55} & \ding{55} & \ding{55} \\
& MetaGraspNet \cite{gilles2022metagraspnet} & 6-DoF & $\sim$12 & 217K & 82 & -& \cmark & \ding{55} & \ding{55} & \ding{55} & \ding{55} & \ding{55} \\
\hline
\multirow{2}{*}{Synth.} 
& Grasp-Anything \cite{graspanything2024} & Rect. & - & 1M & $\sim$3M & $\sim$600M & \cmark & \ding{55} & \ding{55} & \ding{55} & \ding{55} & \ding{55}  \\
& Grasp-Anything++ \cite{lgd2024} & Rect. & - & 1M & $\sim$3M & $\sim$600M & \cmark & \ding{55} & \cmark & \ding{55} & \ding{55} & \ding{55} \\
\hline
\multirow{10}{*}{Real} 
& Cornell \cite{cornell2011}  & Rect. & 1 & 1035 & 240 & 8019 & \ding{55} & \ding{55} & \ding{55} & \ding{55} & \ding{55} & \ding{55}  \\
& Pinto et al. \cite{Pinto2016}  & Rect. & -  & 50K & 150 & 50K & \ding{55} & \ding{55} & \ding{55} & \ding{55} & \ding{55} & \ding{55} \\
& Levine et al. \cite{levine2018learning}  & Rect. & - & 800K & - & 800K& \ding{55} & \ding{55} & \ding{55} & \ding{55} & \ding{55} & \ding{55} \\
& VMRD \cite{vmrd2019}  & Rect. & $\sim$5 & 4683 & $\sim$200 & 100K& \ding{55} & \cmark & \ding{55} & \ding{55} & \ding{55} & \ding{55} \\
& Multi-Object \cite{multiobject2018} & Rect. & $\sim$3 & 96 & - & 2904& \ding{55} & \cmark & \ding{55} & \ding{55} & \ding{55} & \ding{55}\\
& GraspClutter6D \cite{GraspClutter6D2025} &6-DoF & $\sim$15 & 52K & 200 & 9.3B & \cmark & \ding{55} & \ding{55} & \ding{55} & \ding{55} & \ding{55}\\
& OCID-Grasp \cite{ocidgrasp2021}  & Rect. & 1$\sim$20 & 11K & $\sim$40 & 75K & \cmark & \ding{55} & \ding{55} & \ding{55} & \ding{55} & \ding{55}\\
& GraspNet \cite{graspnet2020}& 6-DoF & $\sim$10 & 97K & 88 & $\sim$1.2B& \cmark & \ding{55} & \ding{55} & \ding{55} & \ding{55} & \ding{55} \\
& \textbf{RealVLG-11B (Ours)} & \textbf{Rect.} & \textbf{1$\sim$20} & \textbf{165K} & \textbf{$\sim$800} & \textbf{11B}& \cmark & \cmark & \cmark & \cmark & \cmark & \cmark \\
\hline
\end{tabular}
}
\caption{\textbf{Comparison of RealVLG-11B with existing grasping datasets.} ``–” indicates unknown values.}

\label{tab:grasp_datasets_summary}
\vspace{-8px}
\end{table*}

\section{Related Work}
\label{sec:related_work}

\subsection{Visual-Language Grounding}
In recent years, research on VLG has made significant progress, evolving from task-specific referring expression comprehension \cite{vlg20161, vlg20162} and segmentation \cite{res2016, GRES2023} to open-world grounding empowered by powerful LVLMs \cite{BLIP2023, LLaVA2024, LLAVA2023, bai2025qwen2} . Several studies have introduced multimodal attention mechanisms to predict bounding boxes  \cite{chen2023shikra, CogVLM2025, you2023ferret, liu2025visualrft} or segmentation masks \cite{lisa2024, GLaMM2024, GSVA2024, PSALM2025,YLVLG2025} based on textual semantics, thereby substantially improving object-level semantic localization. Meanwhile, training-free approaches have attracted increasing attention, primarily including CLIP-based methods \cite{subramanian2022reclip, suo2023text, 10204085} and diffusion-based methods \cite{dmseg2023, ni2023ref}, which leverage multimodal representations and diffusion process features, respectively, to achieve zero-shot cross-modal localization.
 Despite these advances, current studies remain limited to coarse object-level grounding and have not been extended to actionable grasp understanding, which involves reasoning about physically grounded object interactions.

\subsection{Robotic Grasping Datasets}

Existing grasping datasets can be broadly classified into simulation-based, synthetic, and real-world categories, each serving different stages of robotic grasp learning. Simulation-based datasets \cite{mahler2017dex, Jacquard2018, VRGrasping2018, Kappler2015, 6dofgraspnet2019, eppner2019billion, egad2020, ACRONYM2021, gilles2022metagraspnet} provide large-scale grasp annotations under controlled environments but lack real-world visual diversity and multimodal semantic grounding. Synthetic datasets (e.g., Grasp-Anything \cite{graspanything2024, lgd2024}) enhance grasp diversity through generative pipelines, yet their semantic grounding remains limited to object-level annotations. Real-world datasets \cite{cornell2011, Pinto2016, levine2018learning, vmrd2019, multiobject2018, GraspClutter6D2025, ocidgrasp2021, graspnet2020} offer more realistic grasp configurations but primarily focus on isolated or cluttered object scenarios, lacking fine-grained semantic alignment and linguistic supervision. In contrast, the proposed RealVLG-11B dataset integrates segmentation masks, bounding boxes, grasp poses, and language descriptions, all verified through both vision-language models and human annotation, ensuring semantic consistency and high labeling reliability.

\subsection{Language-Driven Grasping}

Most existing semantics-driven grasping studies rely on segmentation models \cite{Seamless2019, sam2023, ravi2024sam2} to achieve semantic alignment by isolating the target object from the input scene to assist grasp prediction. Based on this assumption, prior works primarily focus on predicting 4-DoF grasp configurations from RGB-D views \cite{ocidgrasp2021, morrison2018closing, xu2023instance} or 6-DoF grasp poses from 3D data \cite{9197318, qian2024thinkgrasp, tziafas2024towards, vlgrasp2023, graspaffo2019, clipfusion2023}. Although these methods perform well in controlled environments, their heavy reliance on pre-segmented inputs and susceptibility to multi-stage error accumulation hinder their applicability to open-world and language-conditioned manipulation tasks. Recent advances have explored language-guided grasping \cite{meng2024ovgnet, lgd2024, lang6dof2024, tziafas2023language}, where textual instructions are encoded as semantic guidance for grasp generation. However, such methods are typically trained in closed environments, exhibiting limited scalability and weak generalization capabilities. Similarly, task-oriented grasping approaches \cite{murali2021same, GraspGPT2023} leverage the semantic priors of LLMs to enhance task relevance, yet they still overlook key challenges such as visual grounding, reasoning in cluttered scenes, and fine-grained referential understanding, leading to suboptimal semantic consistency. In contrast, our proposed RealVLG-R1 is the first end-to-end robotic perception model built on LVLMs, unifying segmentation, grounding, and grasp perception. Leveraging LVLMs’ multimodal reasoning, it enables integrated grounding and grasp reasoning, reducing multi-stage errors and achieving scalable, semantically consistent robotic perception and manipulation.
\section{RealVLG-11B Dataset}
\label{sec:dataset}

\subsection{Overview}
\begin{figure*}[t]
    \centering
    \includegraphics[width=0.95\textwidth]{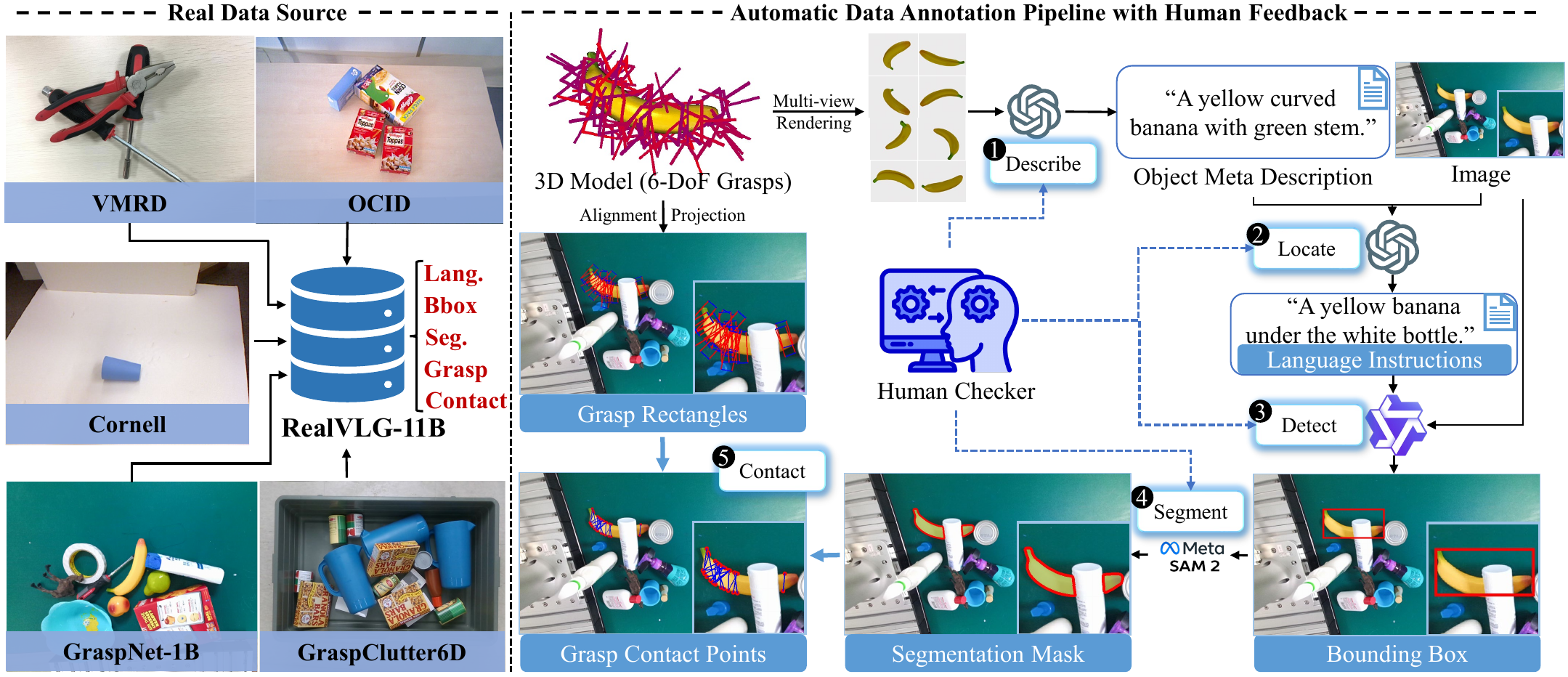}
    \caption{\textbf{Annotation pipeline of the RealVLG-11B dataset.} The pipeline integrates automatic language generation, model-based verification, and manual review to generate high-quality multi-granularity visual and language annotations.}
    \label{fig:realvlg-11b}
    \vspace{-8px}
\end{figure*}

Existing grasping datasets generally suffer from two major limitations. First, most datasets lack a language modality, with annotations primarily focused on geometric and visual aspects in real \cite{cornell2011, Pinto2016, levine2018learning, vmrd2019, multiobject2018, GraspClutter6D2025, ocidgrasp2021, graspnet2020} or simulated \cite{mahler2017dex, Jacquard2018, VRGrasping2018, Kappler2015, 6dofgraspnet2019, eppner2019billion, egad2020, ACRONYM2021, gilles2022metagraspnet} environments, making it difficult to support semantic understanding and task-oriented grasp reasoning. Second, although some datasets \cite{graspanything2024, lgd2024} incorporate language information, they are based on synthetic scenes generated by diffusion model, which can lead to semantic distortion and visual inconsistency. Moreover, their textual descriptions are coarse, preventing precise alignment with specific objects within the scene. To address these issues, as shown in Table \ref{tab:grasp_datasets_summary}, we introduce RealVLG-11B, a large-scale, real-world, multimodal, and multi-granularity visual-language grounding and grasping dataset. Built upon the integration of multiple real-world grasping datasets, RealVLG-11B systematically extends visual annotations and incorporates high-quality natural language descriptions that have been jointly verified by LVLMs and human annotators. These descriptions accurately capture objects and their spatial relationships, enabling semantic-level grasp understanding and reasoning for manipulation. The dataset contains approximately 165,000 images, over 800 object instances, 1.3 million segmentation, detection, and language annotations, and 11 billion grasp examples, providing a high-quality benchmark for multi-granularity perception and manipulation in robotics.

\subsection{Data Source}

AnyGrasp \cite{fang2023anygrasp} has demonstrated the critical importance of real-world data for enabling generalizable robotic geometric grasping in practical applications. Building upon this insight, we have compiled a diverse collection of real-world datasets, encompassing single-object scenes (Cornell \cite{cornell2011}), multi-object relational scenes (VMRD \cite{vmrd2019}), progressively populated object scenes (OCID-Grasp \cite{ocidgrasp2021}), and cluttered environments (GraspNet \cite{graspnet2020}, GraspClutter6D \cite{GraspClutter6D2025}). As summarized in Table \ref{tab:grasp_datasets_summary}, the assembled dataset comprises a total of 165,000 images.

\subsection{Unified Data Annotation}
As shown in Fig. \ref{fig:realvlg-11b}, the construction of RealVLG-11B is predicated on the design of a unified and rigorous data annotation pipeline. This pipeline is essential for integrating diverse real-world grasping datasets and systematically introducing high-precision, multi-granularity visual-language alignment annotations. Crucially, the pipeline not only resolves annotation inconsistencies across disparate data sources (e.g., the presence of only Rectangular Grasp Poses in Cornell \cite{cornell2011} versus their absence in GraspClutter6D \cite{GraspClutter6D2025}) but also ensures the semantic accuracy and visual localizability of the language descriptions through the fusion of LVLMs and detailed human verification.

\textbf{Language Annotation.} The annotation process begins with the generation and refinement of language descriptions. \tikz[baseline=(char.base)]{
    \node[shape=circle, fill=black, text=white, inner sep=1pt] (char) {1};
} For each object in the dataset, its 3D model is extracted and rendered from eight distinct viewpoints, and these multi-view images are input into GPT-4o \cite{openai2024gpt4o} to produce a Meta Description that captures the object’s intrinsic attributes. \tikz[baseline=(char.base)]{
    \node[shape=circle, fill=black, text=white, inner sep=1pt] (char) {2};
} Subsequently, for each target object in every image, both the image and the corresponding Meta Description are provided to GPT-4o, which generates a more detailed Language Instruction describing the object’s category, color, shape, attributes, and, most critically, its spatial relationships with other objects, ensuring that the description uniquely and accurately localizes the target object. 

\textbf{Grounding Annotation.} \tikz[baseline=(char.base)]{
    \node[shape=circle, fill=black, text=white, inner sep=1pt] (char) {3};
} To verify the alignment between the generated Language Description and the object in the image, Qwen-VL-Max \cite{alibaba2025qwenvlmax} is employed as a visual grounding model to detect and output the object’s bounding box based on the image and the language description. \tikz[baseline=(char.base)]{
    \node[shape=circle, fill=black, text=white, inner sep=1pt] (char) {4};
} Finally, to further validate the accuracy of the bounding box, SAM2 \cite{ravi2024sam2} is used to obtain a high-precision segmentation mask for the target object.

\textbf{Grasp Pose Standardization.} In the context of unifying and extending geometric annotations, we focus on generating consistent rectangular grasp poses across all datasets. For datasets such as GraspClutter6D \cite{GraspClutter6D2025}, which lack rectangular grasp annotations but provide 6-DoF grasp poses, we convert these 6-DoF poses into unified Rect Grasp Poses by aligning the object’s 6-DoF pose with its corresponding image instance. \tikz[baseline=(char.base)]{
    \node[shape=circle, fill=black, text=white, inner sep=1pt] (char) {5};
} Based on the resulting Rect Grasp Poses and segmentation masks, grasp contact points are subsequently computed. The computation is designed to ensure that contact points accurately lie on the object surface: if the midpoint along the gripper’s closing direction falls outside the segmentation mask, it is projected onto the mask boundary; if one contact point lies inside the mask while the other lies outside, the external point is shifted along the grasping direction toward the object boundary.

\textbf{Human Verification.} The quality control of the annotation process is maintained through a stringent human-in-the-loop mechanism. Human annotators perform comprehensive cross-validation of the object Meta Description, Language Instruction, Bounding Box detection, and Segmentation Mask to ensure full consistency and accuracy across these four modalities. If any inconsistencies or errors are identified (e.g., imprecise language descriptions or misaligned object localization), the annotators revise the language description and re-execute the bounding box detection and segmentation mask generation procedures, iterating this process until the annotations meet the prescribed high-quality standards.

Based on the annotation pipeline described above, RealVLG-11B achieves a unified integration of multiple real-world grasping datasets, producing high-quality, multi-granularity visual-language annotations. As summarized in Table~\ref{tab:grasp_datasets_summary}, compared to existing datasets, RealVLG-11B provides consistent bounding boxes, segmentation masks,  rectangular grasp poses, and language descriptions that are rigorously verified through both LVLMs and human review. Each object is accompanied by precise language instructions, visually grounded annotations, and grasp contact points, establishing the dataset as the largest and most semantically rich real-world benchmark to date.

\begin{table}
\resizebox{0.45\textwidth}{!}{
    \begin{tabular}{cccccc}
    \hline Set Type & Split & Unique Objects & Images & Instances & Grasps \\
    \hline Training & Train & $\sim 700$ & $\sim 120 \mathrm{~K}$ & $\sim 900 \mathrm{~K}$ & $\sim 10 \mathrm{~B}$ \\
    \hline & Seen & 48 & 15 K & $\sim 150 \mathrm{~K}$ & $\sim 0.3 \mathrm{~B}$ \\
    Evaluation & Similar & 50 & 15 K & $\sim 150 \mathrm{~K}$ & $\sim 0.3 \mathrm{~B}$ \\
    & Novel & 37 & 15 K & $\sim 150 \mathrm{~K}$ & $\sim 0.3 \mathrm{~B}$ \\
    \hline
    \end{tabular}
}
\caption{\textbf{Dataset split for RealVLG-11B.}}
\label{tab:dataset_split}
\vspace{-18px}
\end{table}

\subsection{Evaluation}

To evaluate open-world generalization, as shown in Table \ref{tab:dataset_split}, we split $\sim$800 RealVLG-11B objects into three subsets: Seen (48 training objects), Similar (50 similar but unseen objects), and Novel (37 novel objects). Each subset contains $\sim$15K images, $\sim$150K object instances, and 300M grasp annotations, providing a large and diverse benchmark for visual-language grounding and grasping reasoning.

\section{RealVLG-R1 Model}
\label{sec:method}

\subsection{Overview}
\begin{figure*}[t]
    \centering
    \includegraphics[width=0.99\textwidth]{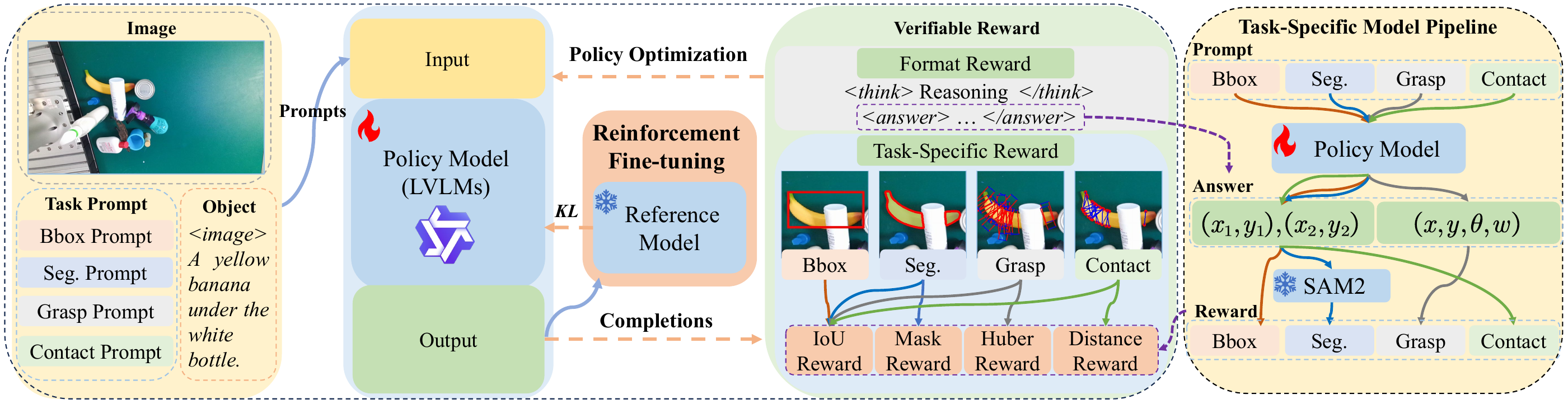}
   \caption{\textbf{Framework of RealVLG-R1.} RealVLG-R1 fine-tunes pretrained LVLMs via reward-driven RL using task-specific verifiable rewards, enabling adaptive learning and improved generalization over bounding boxes, segmentation, grasp rectangles, and contact points.}
    \label{fig:realvlg-r1}
    \vspace{-8px}
\end{figure*}

Existing LVLMs \cite{bai2025qwen2, LLaVA2024} demonstrate strong generalization capabilities in visual understanding and language instruction following. However, extending these models to Visual-Language Grounding and Grasping tasks remains non-trivial. Traditional Supervised Fine-Tuning (SFT) methods \cite{radford2019language, brown2020gpt3, ouyang2022training} perform effectively on tasks with explicit and deterministic supervision, such as object detection \cite{xiao2024florence, chen2021pix2seq}. In contrast, grasp pose prediction inherently involves multiple feasible solutions, as a single object can afford multiple stable grasp configurations or contact points. This inherent multiplicity introduces two major challenges for SFT:
(1) enforcing the model to fit a single label often yields ``averaged” predictions that are physically infeasible; and
(2) fixed supervision signals are insufficient to capture multi-dimensional objectives such as grasp stability, spatial consistency, and physical plausibility.

To address these challenges, as shown in Fig. \ref{fig:realvlg-r1}, we propose a unified framework, RealVLG-R1, which fine-tunes pretrained LVLMs using a reinforcement-style optimization strategy inspired by DeepSeek-R1 \cite{guo2025deepseek}. This strategy draws on the principles of Reinforcement Learning, updating model policies based on reward-driven feedback rather than fixed ground-truth annotations. Furthermore, we introduce a Verifiable Reward Mechanism that dynamically evaluates and guides model predictions in terms of both semantic correctness and physical feasibility. By relaxing the dependence on deterministic labels, RealVLG-R1 enables adaptive optimization in multi-solution scenarios, effectively bridging visual-language understanding with physical interaction, and promoting the generalization of language-driven grounding and grasping to real-world tasks.

\subsection{Policy Optimization with Verifiable Rewards}
The policy fine-tuning of RealVLG-R1 adopts the Reinforcement Learning with Verifiable Rewards (RLVR) paradigm \cite{guo2025deepseek, liu2025visualrft, team2025kimi}, aiming to align the pretrained LVLM policy $\pi_\theta$ with precise visual grounding and grasping  tasks. Unlike conventional approaches \cite{liu2024skywork, brown2020gpt3, RLHFV2024} that rely on learned reward models, RLVR leverages task-intrinsic verifiable criteria to provide objective reward signals through a binary verification function $R(q, o)$. Formally, the verifiable reward function $R(q, o)$ is defined as,
\begin{equation}
R(q, o) =
\begin{cases}
1, & \text{if } o = \text{ground truth}, \\
0, & \text{otherwise}.
\end{cases}
\end{equation}
where $q$ denotes the input query and $o$ represents the response generated by the policy model $\pi_\theta$.
Furthermore, the objective of RealVLG-R1 aims to maximize the expected reward while introducing a KL-divergence regularization term $\mathbb{D}_\text{KL}$ \cite{schulman2017proximal}, controlled by the hyperparameter $\beta$, to constrain policy updates. This regularization ensures that the optimized policy $\pi_\theta$ remains close to the reference model $\pi_{\text{ref}}$ and prevents excessive deviation during fine-tuning:
\begin{equation}
\max_{\pi_\theta} \mathbb{E}_{q, o} 
\Big[ R(q, o) - \beta \, \mathbb{D}_\text{KL}[\pi_\theta(o|q) \parallel \pi_{\text{ref}}(o|q)] \Big].
\end{equation}

To eliminate dependence on additional value models during policy optimization, 
RealVLG-R1 employs  a group-based contrastive advantage estimation strategy. 
Specifically, $G$ responses $\{y_i\}_{i=1}^G$ are sampled from the old policy $\pi_{\theta_{\text{old}}}$, and their verifiable rewards $r(x, y_i)$ are used to compute a group-relative advantage $\widehat{A}_i$, which is shared across all tokens within the sequence:
\begin{equation}
\widehat{A}_i = \frac{r(x, y_i) - \mathrm{mean}(\{r(x, y_j)\}_{j=1}^G)}
{\mathrm{std}(\{r(x, y_j)\}_{j=1}^G)}.
\end{equation}

Initially, RealVLG-R1 employs the GRPO \cite{shao2024grpo} algorithm, which applies token-level importance weights $w_{i,t}(\theta)$ for off-policy correction:
\begin{equation}
\small
\begin{aligned}
\mathcal{L}_{\text{GRPO}}(\theta) &= 
- \mathbb{E}_{x, \{y_i\}} \Bigg[ 
\frac{1}{G} \sum_{i=1}^{G} \frac{1}{|y_i|} 
\sum_{t=1}^{|y_i|} \Big[ 
\min \Big( w_{i,t}(\theta) \widehat{A}_i, \\
& \mathrm{clip} \big( w_{i,t}(\theta) , 1 - \varepsilon, 1 + \varepsilon \big) \widehat{A}_i \Big) 
- \beta \, \mathbb{D}_{\text{KL}} 
\Big] 
\Bigg],
\end{aligned}
\end{equation}
where the token-level weight $w_{i,t}(\theta)$ is defined as:
\begin{equation}
    w_{i,t}(\theta) = \frac{\pi_\theta}{\pi_{\theta_{\text{old}}}} = 
\frac{\pi_\theta(y_{i,t}|x, y_{i,<t})}{\pi_{\theta_{\text{old}}}(y_{i,t}|x, y_{i,<t})}.
\end{equation}

Furthermore, to mitigate the high variance and instability induced by token-level weights in long-sequence tasks, RealVLG-R1 adopts the GSPO \cite{zheng2025group} approach. GSPO introduces clipped importance weights $s_i(\theta)$ at the sequence level, thereby enhancing robustness in scenarios involving long outputs. The GSPO loss is formulated as:
\begin{equation}
\begin{aligned}
\mathcal{L}_{\text{GSPO}}(\theta) &= 
- \mathbb{E}_{x, \{y_i\}} \Bigg[ 
\frac{1}{G} \sum_{i=1}^{G} \Big[ 
\min \big( s_i(\theta) \widehat{A}_i, \\
& \mathrm{clip}(s_i(\theta), 1 - \varepsilon, 1 + \varepsilon) \widehat{A}_i \big) 
- \beta \, \mathbb{D}_{\text{KL}} 
\Big] 
\Bigg],
\end{aligned}
\end{equation}
where sequence-level weights are length-normalized to control variance:
\begin{equation}
s_i(\theta) = \Big( \frac{\pi_\theta(y_i|x)}{\pi_{\theta_{\text{old}}}(y_i|x)} \Big)^{\frac{1}{|y_i|}}.
\end{equation}

\subsection{Task-Specific Pipelines and Verifiable Rewards}
To achieve a unified model pipeline for visual-language grounding and grasping tasks, supporting coarse-to-fine perception capabilities, RealVLG-R1 constructs task-specific pipelines and designs composite verifiable reward functions tailored for various robotic visual tasks. 

As shown in Fig. \ref{fig:realvlg-r1}, the policy model receives an image and a task prompt, then generates structured outputs according to task requirements. All tasks adopt the \texttt{<think>...</think>\textbackslash n<answer>...</answer>} format to facilitate the computation of format rewards. Within the \texttt{<answer>} component, detection (Bbox), segmentation (Seg.), and contact point (Contact) prediction tasks output two point coordinates $(x_1, y_1), (x_2, y_2)$, whereas grasp pose (Grasp) prediction tasks output a grasp pose $(x, y, \theta, w)$. Notably, for segmentation tasks, the model first predicts a bounding box, which is then passed to a frozen SAM2 \cite{ravi2024sam2} model to generate the final mask.

The core of RealVLG-R1 is its composite reward function $R(q, o)$, providing hierarchical and verifiable feedback by combining output format compliance with task-specific geometric accuracy:
\begin{equation}
    R(q, o) = R_{\text{Format}} + R_{\text{Task}},
\end{equation}
where $R_{\text{Format}}$ enforces structured outputs, and $R_{\text{Task}}$ is tailored to four downstream visual-language tasks, optimizing the alignment between predictions and ground truth.

\textbf{Bbox.} For a predicted bounding box $B_p$ and ground truth $B_{gt}$, a binary IoU-based reward is used:
\begin{equation}
R_{\text{Bbox}} = \mathbf{1}(\text{IoU}(B_p, B_{gt}) \ge \tau_{\text{iou}}),
\end{equation}
where $\tau_{\text{iou}}$ is a predefined IoU threshold ensuring minimum coarse-grained localization accuracy.

\textbf{Seg.}  
For segmentation, the reward combines coarse localization and fine-grained mask accuracy:
\begin{equation}
R_{\text{Seg}} = \mathbf{1}(\text{IoU}(B_p, B_{gt}) \ge \tau_{\text{iou}}) + \text{S}_{\alpha}(M_p, M_{gt}),
\end{equation}
where $M_p$ and $M_{gt}$ are the predicted and ground-truth masks, and $\text{S}_{\alpha}$ is the S-measure \cite{smeasure2017}.

\textbf{Grasp.} Robotic grasping requires the accurate prediction of a 4D pose $G_p = (x_p, y_p, \theta_p, w_p)$. To handle the periodicity of $\theta$ and enable smooth policy updates, the angle is represented using its sine and cosine components. Then, the grasping reward is formulated as the negative sum of Huber losses computed over all pose components:
\begin{equation}
    R_{\text{Grasp}} = - \sum_{v \in \{x, y, \cos\theta, \sin\theta, w\}} 
\mathcal{L}_{\text{Huber}}(v_p, v_{gt}),
\end{equation}
where $\mathcal{L}_{\text{Huber}}$ is the Huber loss applied per component. 

\textbf{Contact.} The reward for contact point prediction considers both rectangle alignment and point accuracy. The predicted contact points $P_1^p, P_2^p$ are first converted into a rectangular grasp pose $G_p$ with fixed width, and then the rectangle alignment and point accuracy are evaluated by:
\begin{equation}
\small
R_{\text{Contact}} = \mathbf{1}(\text{IoU}(G_p, G_{gt}) \ge \tau_{\text{iou}}) 
- \sum_{i=1}^{2} \| P_i^p - P_i^{gt} \|_2.
\end{equation}

Based on the aforementioned design, RealVLG-R1 achieves unified perception across multiple visual-language tasks, encompassing both coarse-grained localization and fine-grained reasoning, thereby exemplifying a unified approach to vision-language grounding and robotic interaction. By incorporating verifiable supervision, the model effectively bridges perception and action, enabling robust zero-shot deployment in real-world environments.
\section{Experiments}
\label{sec:experiments}
\subsection{Data Quality Evaluation}
To comprehensively evaluate the proposed RealVLG-11B dataset in terms of linguistic diversity and visual annotation quality, we conducted a comparative analysis against Grasp-Anything~\cite{graspanything2024} and Grasp-Anything++~\cite{lgd2024}. A total of 10,000 samples were randomly selected from each dataset for evaluation. The assessment employed several metrics: linguistic diversity measured by MTLD \cite{McCarthy2010MTLD}, visual-language alignment quantified by the CLIP Score ($\text{S}_{\text{CLIP}}$) \cite{radford2021learning}, and three spatial consistency indicators: segmentation coverage ratio of bounding boxes ($\text{R}_s$), proportion of grasp points located within segmentation masks ($\text{R}_g$), and proportion of contact centers falling inside segmentation masks ($\text{R}_c$). It is noteworthy that in the Grasp-Anything datasets, bounding boxes are directly derived from segmentation masks, rendering its $\text{R}_s$ metric inapplicable.

As shown in Table~\ref{tab:realvlg_metrics}, benefiting from our carefully designed LVLM-assisted and human double-review annotation pipeline, RealVLG-11B consistently outperforms existing datasets across all comparable metrics. Specifically, since the language descriptions in RealVLG-11B encompass not only object categories but also rich attributes and spatial relations, it achieves a higher MTLD score, reflecting greater linguistic richness. The CLIP Score and our LVLM-based bounding box annotation mechanism further demonstrate that RealVLG-11B achieves precise semantic alignment between textual descriptions and visual objects, exhibiting strong visual-language consistency. In terms of spatial consistency, RealVLG-11B attains a near-perfect Seg-in-Bbox ($\text{R}_s$) of 0.99, validating the high geometric coherence between bounding boxes and segmentation masks. Unlike Grasp-Anything, which relies on low-quality grasp poses generated by RAGT-3/3 \cite{cao2023nbmod}, RealVLG-11B inherits meticulously curated grasp annotations from prior grasping datasets, leading to a higher Grasp-in-Seg score ($\text{R}_g$, $\sim$0.3 $\uparrow$) and better alignment with language descriptions. Overall, RealVLG-11B demonstrates superior performance in textual richness, multimodal alignment, and annotation reliability, establishing a high-quality foundation for future research on visual-language grounding and grasping.

\begin{table}[t]
\centering
\resizebox{0.46\textwidth}{!}{
\begin{tabular}{lccccc}
    \hline
    Dataset & MTLD $\uparrow$ & $\text{S}_{\text{CLIP}}$ $\uparrow$ & $\text{R}_s$ $\uparrow$ & $\text{R}_g$ $\uparrow$ & $\text{R}_c$ $\uparrow$ \\
    \hline
    Grasp-Anything~\cite{graspanything2024} & 27.45 & 0.54 & -- & 0.38 & 0.69 \\
    Grasp-Anything++~\cite{lgd2024} & 15.14 & 0.52 & -- & 0.31 & 0.62 \\
    RealVLG-11B (Ours) & \textbf{36.49} & \textbf{0.65} & \textbf{0.99} & \textbf{0.69} & \textbf{0.87} \\
    \hline
\end{tabular}
}
\caption{ \textbf{Linguistic and grounding quality comparison.}}
\label{tab:realvlg_metrics}
\vspace{-12px}
\end{table}

\begin{table*}[t]
\centering
\resizebox{0.98\textwidth}{!}{
\begin{tabular}{l|cccc|cccc|cccc}
\toprule
\multirow{2}{*}{Method} 
& \multicolumn{4}{c|}{Seen} 
& \multicolumn{4}{c|}{Similar} 
& \multicolumn{4}{c}{Novel} \\
\cmidrule(lr){2-5} \cmidrule(lr){6-9} \cmidrule(lr){10-13}
& Bbox & Seg. & Grasp & Contact 
& Bbox & Seg. & Grasp & Contact 
& Bbox & Seg. & Grasp & Contact \\
& gIoU/cIoU/R$_v$ & F$_\beta$/S$_\alpha$/R$_v$ & mIoU/gAcc/R$_v$ & mIoU/gAcc/R$_v$
& gIoU/cIoU/R$_v$ & F$_\beta$/S$_\alpha$/R$_v$ & mIoU/gAcc/R$_v$ & mIoU/gAcc/R$_v$
& gIoU/cIoU/R$_v$ & F$_\beta$/S$_\alpha$/R$_v$ & mIoU/gAcc/R$_v$ & mIoU/gAcc/R$_v$ \\
\midrule
Qwen-VL-Max \cite{alibaba2025qwenvlmax} 
& \textbf{92.3}/\textbf{93.2}/61.3 & \textbf{87.8}/\textbf{44.8}/71.5 & \textbf{16.0}/\textbf{16.7}/14.2 & \textbf{34.7}/\textbf{25.6}/15.4 
& \textbf{89.4}/\textbf{90.0}/64.3 & \textbf{86.8}/\textbf{43.5}/66.4 & \textbf{14.9}/\textbf{7.1}/17.9 & \textbf{32.0}/\textbf{7.0}/18.3 
& \textbf{88.4}/\textbf{88.8}/67.7 & \textbf{47.3}/\textbf{38.6}/64.0 & \textbf{8.1}/\textbf{5.4}/22.6 & \textbf{18.8}/\textbf{20.0}/18.3 \\

Gemini2.5-Flash \cite{deepmind_gemini_flash} 
& 60.1/61.5/\textbf{83.4} & 3.3/14.1/\textbf{83.4} & 3.2/1.2/\textbf{100} & 5.2/3.6/\textbf{100}
& 59.2/60.5/\textbf{80.9} & 3.8/13.1/\textbf{80.9} & 2.5/2.2/\textbf{100} & 6.3/4.7/\textbf{100}
& 62.6/63.7/\textbf{95.1} & 3.9/14.9/\textbf{95.1} & 3.4/1.9/\textbf{100} & 4.0/4.3/\textbf{100}  \\
\midrule

Qwen2.5VL-3B \cite{bai2025qwen2} 
& 50.0/51.9/99.2 & 75.1/43.5/99.2 & 4.3/2.9/94.1 & 25.2/17.7/87.0 
& 51.2/52.4/94.5 & 75.1/42.4/94.5 & 4.0/1.4/92.3 & 27.8/22.9/85.5 
& 52.9/53.7/96.9 & 45.9/38.7/97.0 & 3.1/1.3/95.7 & 14.5/8.6/98.8 \\
Qwen2.5VL-3B + SFT 
& 56.4/57.1/100 & 76.2/43.4/100 & 3.4/1.7/100 & 28.6/21.4/100 
& 54.7/55.9/100 & 75.4/43.2/100 & 3.6/2.1/100 & 29.3/28.2/100
& 57.2/58.6/100 & 46.2/38.9/100 & 4.4/1.5/100 & 19.7/15.8/100 \\
RealVLG-R1-3B (GRPO) 
& 87.2/87.6/96.1 & 87.4/45.2/96.0 & \textbf{34.7}/\textbf{40.3}/\textbf{100} & \textbf{51.9}/\textbf{33.2}/\textbf{100}
& \textbf{86.5}/\textbf{87.1}/97.9 & \textbf{85.8}/\textbf{43.9}/97.9 & \textbf{28.2}/\textbf{31.9}/\textbf{100} & \textbf{47.8}/\textbf{31.9}/\textbf{100} 
& \textbf{78.5}/\textbf{79.5}/98.2 & \textbf{50.3}/39.4/98.2 & \textbf{16.3}/\textbf{17.1}/\textbf{100} & \textbf{26.9}/\textbf{20.2}/\textbf{100} \\
RealVLG-R1-3B (GSPO) 
& \textbf{87.7}/\textbf{88.4}/\textbf{100} & \textbf{88.7}/\textbf{45.3}/\textbf{100} & 29.2/33.6/100 & 48.6/29.6/100 
& 84.6/85.3/\textbf{100} & 85.0/43.7/\textbf{100} & 27.6/30.6/100 & 46.5/28.1/100 
& 78.0/79.3/\textbf{100} & 49.1/\textbf{39.6}/\textbf{100} & 15.3/9.1/100 & 26.9/17.7/100 \\

\midrule
Qwen2.5VL-7B \cite{bai2025qwen2} 
& 56.8/58.3/100  & 80.6/43.8/100 & 4.1/1.8/43.5  & 33.7/26.3/99.2 
& 57.3/57.9/99.6 & 78.0/42.4/99.6 & 3.2/0.0/49.4 & 32.8/23.5/99.6
& 55.6/56.8/100 & 48.3/39.2/100 & 1.6/0.0/64.6 & 20.0/12.8/100 \\
Qwen2.5VL-7B + SFT 
& 64.8/63.5/100 & 82.2/44.1/100 & 5.1/3.1/80.6 & 38.7/31.2/100 
& 66.7/62.7/100 & 79.6/42.6/100 & 4.6/2.7/85.4 & 37.4/29.6/100 
& 64.9/65.2/100 & 48.5/39.4/100 & 3.5/1.4/78.1 & 26.4/16.4/100 \\
RealVLG-R1-7B (GRPO) 
& 88.0/88.5/100 & \textbf{89.2}/45.3/100 & 32.2/\textbf{39.3}/99.6 & 53.3/33.1/99.2
& 83.4/84.2/100 & 85.0/43.8/100 & \textbf{28.8}/\textbf{36.3}/99.6 & 50.5/34.9/97.4
& 84.6/85.4/100 & \textbf{52.2}/\textbf{39.7}/\textbf{100} & \textbf{17.0}/\textbf{19.1}/98.8 & \textbf{30.1}/20.7/100 \\
RealVLG-R1-7B (GSPO) 
& \textbf{89.0}/\textbf{89.6}/\textbf{100} & 88.9/\textbf{45.4}/\textbf{100} & \textbf{33.6}/32.8/\textbf{100} & \textbf{55.3}/\textbf{37.2}/\textbf{100} 
& \textbf{86.4}/\textbf{86.9}/\textbf{100} & \textbf{86.6}/\textbf{43.8}/\textbf{100} & 27.9/30.2/\textbf{100} & \textbf{50.6}/\textbf{36.2}/\textbf{100} 
& \textbf{88.5}/\textbf{88.9}/\textbf{100} & 52.0/39.7/\textbf{100} & 16.5/18.3/\textbf{100} & 28.9/\textbf{25.0}/\textbf{100} \\
\bottomrule
\end{tabular}
}
\caption{
\textbf{RealVLG benchmark comprehensive results.} All metrics are reported in percentage format. 
}
\label{tab:realvlg_benchmark}
\vspace{-12px}
\end{table*}

\subsection{RealVLG Benchmark}
\paragraph{Evaluation Metrics.}
To systematically evaluate the multi-task performance on the RealVLG benchmark, we utilize a task-specific evaluation framework that rigorously addresses both geometric precision and semantic consistency. For object localization, we quantify bounding box spatial alignment using Generalized IoU (gIoU) \cite{giou2019} and Complete IoU (cIoU) \cite{zheng2021ciou}. Segmentation quality is assessed via the S-measure ($\text{S}_\alpha$) \cite{smeasure2017} for structural similarity and the F-measure ($\text{F}_\beta$) \cite{fmeasure2023} to balance pixel-level precision and recall. In rectangular grasp pose prediction, performance relies on mean IoU (mIoU) and Grasp Accuracy (gAcc) \cite{cornell2011}, where gAcc is achieved when the IoU exceeds 0.25 and the angular deviation is below $30^\circ$. Furthermore, contact point predictions are standardized into rectangular grasps for unified evaluation with the main grasping task. Importantly, considering the potential for output instability in LVLMs, we introduce a Validity Rate (VR, R$_v$) metric to quantify the proportion of valid outputs, defined as predictions that are non-empty and structurally parsable. The aforementioned accuracy metrics are calculated exclusively over this subset of successful samples. This comprehensive and integrated methodology establishes a unified, quantifiable standard for comparing model architectures across visual-language grounding and grasping tasks.

\textbf{Baselines.}
To systematically evaluate model performance on visual-language grounding and grasping tasks, the RealVLG-Benchmark considers two categories of baseline models for comparison: closed-source models (Qwen-VL-Max \cite{alibaba2025qwenvlmax} and Gemini2.5-Flash \cite{deepmind_gemini_flash}) and open-source models (the Qwen2.5-VL series \cite{bai2025qwen2}). Notably, Gemini2.5-Flash can directly produce segmentation outputs, whereas Qwen-VL-Max achieves segmentation via bounding box predictions combined with SAM2 \cite{ravi2024sam2} mask generation. For the open-source baselines, we evaluated Qwen2.5-VL and its supervised fine-tuned variant implemented under the LLaMA-Factory \cite{zheng2024llamafactory} framework (Qwen2.5-VL+SFT). Building upon this, our proposed RealVLG-R1 model employs Qwen2.5-VL as its backbone and is developed within the VERL framework \cite{verl2025}. Compared to traditional SFT models, the key advantage of RealVLG-R1 lies in its ability to achieve efficient training with only a small amount of high-quality data. Following the data split in Table \ref{tab:dataset_split}, both RealVLG-R1 and Qwen2.5-VL+SFT were fine-tuned for 10 epochs using only 10\% of the training set.

\textbf{Experimental Analysis.}
According to the quantitative results in Table~\ref{tab:realvlg_benchmark}, the experimental outcomes can be systematically analyzed from two perspectives. 

(1) Visual Grounding. The closed-source model Qwen-VL-Max achieves high localization accuracy (gIoU/cIoU of 92.3\%/93.2\%) and strong segmentation consistency (F$_\beta$=87.8\%, S$_\alpha$=44.8\%), benefiting from extensive pretraining on large-scale vision–language data. However, its execution reliability (R$_v$) remains around 60–70\%, indicating strong visual understanding but limited structured response capability. Among open-source models, Qwen2.5-VL exhibits considerably lower localization and segmentation accuracy (gIoU$\sim$50\%, F$_\beta$$\sim$75\%) than closed-source counterparts but maintains nearly perfect output consistency (R$_v$$\sim$100\%), suggesting that while its structured generation is reliable, its semantic grounding remains limited. SFT provides only marginal improvements ($\sim$5\% gain in gIoU), enhancing stability rather than geometric precision. In contrast, RealVLG-R1 delivers substantial gains across all settings: the 3B model achieves a gIoU exceeding 87\% (a 30-point gain over SFT), and the 7B version further rises to 89\%, maintaining strong generalization in Novel scenes (gIoU$\sim$88\%). All configurations achieve 100\% R$_v$, indicating that reinforcement learning markedly enhances both interpretability and consistency.

(2) Visual Grasping. The visual grasping task imposes higher demands on physical reasoning and action coherence. Owing to the absence of task-specific pretraining data, Qwen2.5-VL and its SFT variant perform poorly (mIoU$<$5\%, gAcc$<$3\%), suggesting that token-level supervision improves output formatting but not physical plausibility. Reinforcement learning substantially boosts performance: for the 3B model, mIoU/gAcc reach 34.7\%/40.3\% on the Seen set, improving by over 30 points compared to SFT. The GSPO variant attains slightly lower geometric scores (mIoU=29.2\%) but exhibits higher training stability and fully structured outputs (R$_v$=100\%). Even in Novel scenarios, RealVLG-R1 sustains reasonable grasp and contact accuracy (mIoU/gAcc=26.9\%/20.2\%), reflecting nontrivial generalization. The 7B model follows a similar pattern that GRPO yields slightly higher grasping precision, whereas GSPO achieves superior stability and contact accuracy (mIoU/gAcc=55.3\%/37.2\%). These results suggest that GRPO’s reward formulation better enhances fine-grained action precision in smaller models, while GSPO’s sequence-level incentives provide smoother optimization for larger models, albeit at the cost of diluted reward sensitivity in small-scale settings. 
\section{Conclusions}
\label{sec:conclusions}
We introduced RealVLG, consisting of the RealVLG-11B dataset and RealVLG-R1 model, to unify vision-language grounding and grasping. It predicts bounding boxes, segmentation masks, grasp poses, and contact points from natural language with strong zero-shot generalization, providing a benchmark and platform for language-driven robotic perception. Future work will extend RealVLG to 3D space, and explore efficient models such as SmolVLM \cite{marafioti2025smolvlm} to improve runtime without extra fine-tuning.

\section*{Acknowledgments}
This work was supported in part by the New Generation Artificial Intelligence-National Science and Technology Major Project under Grant 2025ZD0123701, in part by the National Natural Science Foundation of China under Grants 62272343 and 62476202, and in part by the Fundamental Research Funds for the Central Universities.

{
    \small
    \bibliographystyle{ieeenat_fullname}
    \bibliography{main}
}

\clearpage
\setcounter{page}{1}

\maketitlesupplementary

\begin{strip}
\begin{center}
    \includegraphics[width=0.98\linewidth]{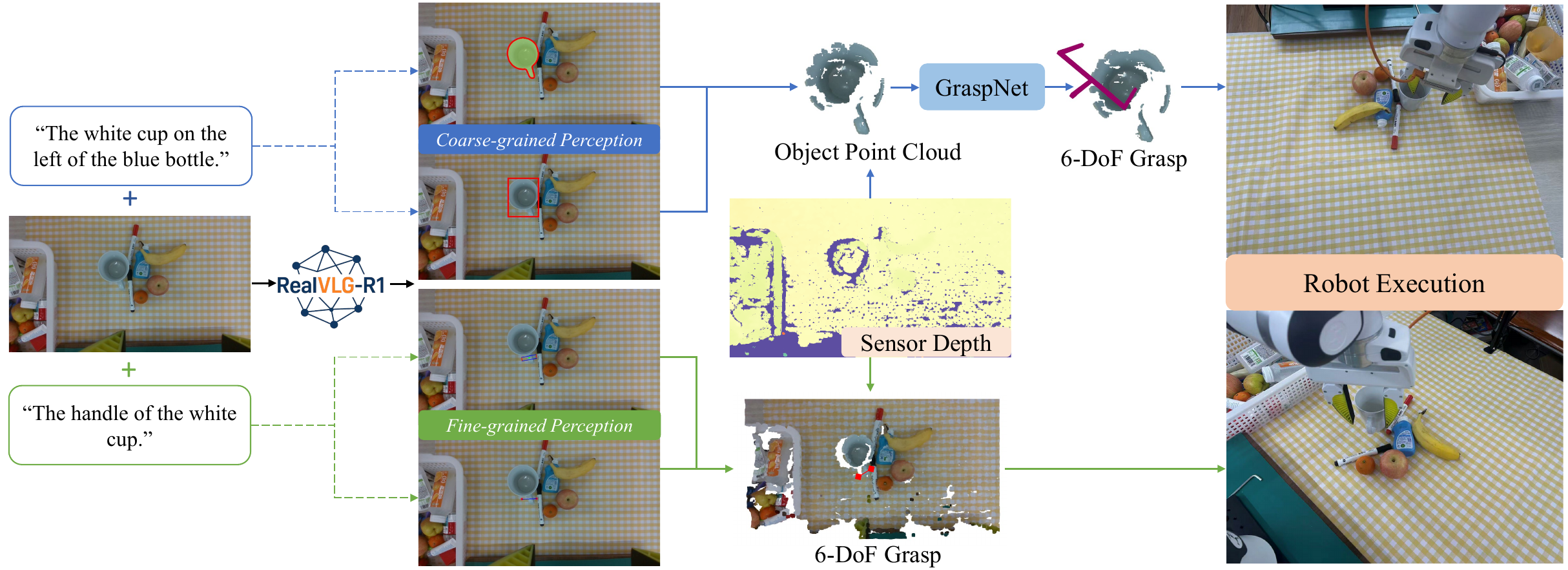}
    \captionof{figure}{\textbf{Overview of the RealVLG-R1 deployment for real-world Visual-Language Grasping tasks.} 
    RealVLG-R1 produces multi-granularity visual–language outputs, which can be leveraged in two complementary grasping strategies: 
    (a) coarse-grained, object-centric grasping, where segmentation masks or bounding boxes are projected into 3D point clouds to generate 6-DoF grasp poses via a 3D grasping module; 
    (b) fine-grained, part-level grasping, where 2D grasp predictions are directly transformed into executable 6-DoF poses using depth and camera parameters, enabling semantically precise manipulation. 
    This design supports hierarchical control from global geometry to detailed semantic structures.}
    \label{fig:realvlg_deploy}
\end{center}
\end{strip}

\section{Real-world Visual-Language Grasping}
\label{sec:suppl-realvlg-r1}
As illustrated in Fig. \ref{fig:realvlg_deploy}, this section describes how the multi-granularity visual–language understanding capabilities of RealVLG-R1 are deployed in real-world Visual-Language Grasping tasks. To address different robotic application requirements, we design two complementary grasping strategies: a coarse-grained, object-centric approach and a fine-grained, part-level approach, enabling hierarchical control from global geometry to detailed semantic structures.

\textbf{Coarse-Grained Grasping (Object-Level).}
In the coarse-grained strategy, RealVLG-R1 outputs object bounding boxes (Bbox) or segmentation masks (Seg.), which are used to generate a binary mask $\mathbf{M}$. The corresponding depth region is first extracted from the sensor depth $\mathbf{D}$:
\begin{equation}
    \mathbf{D}_{\text{obj}} = \mathbf{D} \otimes \mathbf{M},
\end{equation}
and then projected into 3D space using the camera intrinsic and extrinsic parameters :
\begin{equation}
    \hat{P}_{\text{obj}} = \pi^{-1}(\mathbf{D}_{\text{obj}}, K, T),
\end{equation}
where $\pi^{-1}(\cdot)$ denotes the standard depth-to-point-cloud back-projection operation. The resulting object point cloud $\hat{P}_{\text{obj}}$ is subsequently input to a 3D grasping model such as GraspNet \cite{graspnet2020} to generate multiple candidate 6-DoF grasp poses. A grasp planner then selects the optimal pose, which is executed by the robotic manipulator. This pipeline establishes a staged process of ``2D visual–language localization → point cloud reconstruction → 6-DoF grasp generation,” demonstrating stability in conventional object grasping scenarios. However, grasp quality is constrained by depth noise and occlusions, and the generated grasp poses typically lack fine-grained control over specific semantic parts.

\textbf{Fine-Grained Grasping (Part-Level).}
In the fine-grained strategy, RealVLG-R1 directly predicts 2D grasp priors on the image plane, including rectangular grasp poses or grasp contact points. To convert these 2D predictions into executable 6-DoF poses, each pixel $(u,v)$ is first projected into the camera coordinate system using the corresponding sensor depth value $d$ and camera intrinsics $K$:
\begin{equation}
    \mathbf{p}_c = d \cdot K^{-1} 
    \begin{bmatrix} u & v & 1 \end{bmatrix}^\top
\end{equation}
For a predicted grasp rectangle $\mathcal{B} = \{(u_i,v_i)\}_{i=1}^4$, the four corner points are projected to obtain $\{\mathbf{p}_c^{(i)}\}_{i=1}^4$, from which the grasp direction and normal vector are computed to define a local grasp coordinate frame $R_c$ and center position $\mathbf{t}_c$. Finally, the local grasp pose is transformed into the robot coordinate system using the camera-to-robot extrinsics $[R_{cam}^{rob} | \mathbf{t}_{cam}^{rob}]$:
\begin{equation}
    \mathbf{t}_{rob} = R_{cam}^{rob}\mathbf{t}_c + \mathbf{t}_{cam}^{rob}, \quad
    R_{rob} = R_{cam}^{rob} R_c.
\end{equation}
This procedure establishes a direct mapping from 2D model predictions to executable 6-DoF grasp poses, enabling semantically precise part-level grasping, such as ``grasping the handle of a cup”, while maintaining both grasping efficiency and semantic consistency.

Overall, the two strategies offer complementary advantages: coarse-grained grasping provides a stable and generalizable approach with clear staged execution but limited semantic specificity, whereas fine-grained grasping enables semantically aligned, part-level manipulation but imposes higher requirements on geometric accuracy and depth estimation. In future work, we aim to extend RealVLG-R1 to full 3D Visual-Language Grounding and Grasping, allowing the system to generate geometrically accurate and semantically consistent executable grasp poses directly in 3D space, thereby enabling more robust and precise language-driven robotic manipulation.

\section{Details of RealVLG}
\subsection{Details of RealVLG-11B Dataset}
\begin{figure*}[t]
    \centering
    \includegraphics[width=0.98\textwidth]{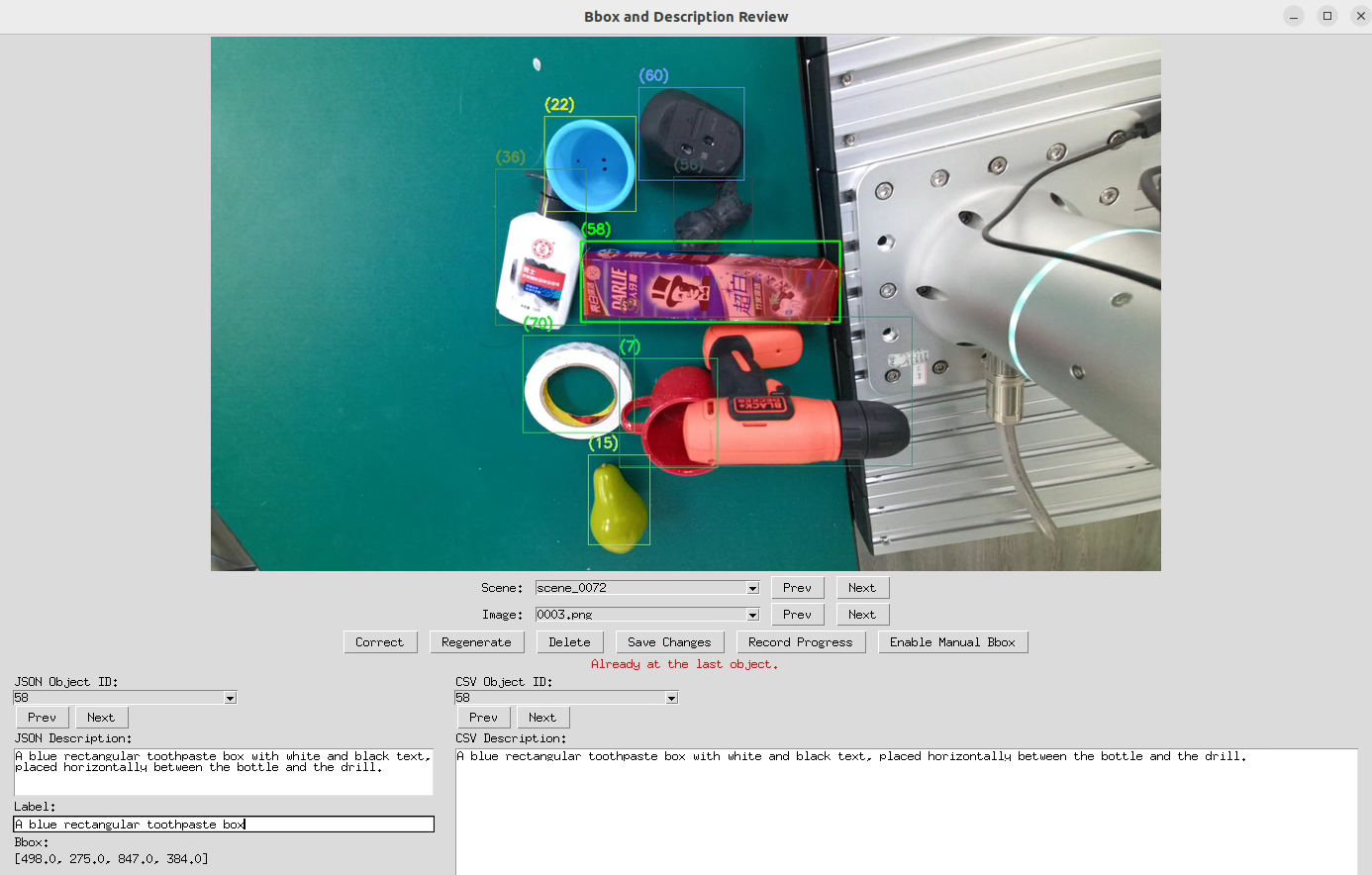}
    \caption{\textbf{
    Human-Verification System.} 
    This application provides an interactive interface for human-in-the-loop verification, allowing users to review, correct, and confirm automatically generated visual-language annotations. It serves as a crucial component for ensuring the quality and reliability of RealVLG-11B dataset annotations.}
    \label{fig:realvlg-human}
\end{figure*}

In Fig. \ref{fig:realvlg-11b} of the main text, the Object Meta Description and Localization Description of each object are defined by Prompt \ref{prompt:object_meta_prompt} and Prompt \ref{prompt:location_prompt}, respectively, and are fed into GPT-4o \cite{openai2024gpt4o} to generate diverse yet semantically consistent linguistic descriptions. The Detection Prompt (Prompt \ref{prompt:detection_prompt}) is subsequently input to Qwen-VL-Max \cite{alibaba2025qwenvlmax} to predict the bounding boxes of the target objects, thereby achieving structured alignment between linguistic and visual modalities. Furthermore, as illustrated in Fig. \ref{fig:realvlg-human}, the human verification phase involves annotators reviewing and refining the automatically generated results via the Human-Verification System, ensuring both accuracy and consistency of the annotations. The entire annotation process was collaboratively conducted by three annotators over a period of approximately five months, ultimately yielding a high-quality, multi-granularity vision-language annotation dataset.

\begin{promptbox}{Object Metadata Description Prompt}
\label{prompt:object_meta_prompt}

Look at the provided images of the object \texttt{"{obj\_name}"} and write a concise English description, 
including its color, shape, and category. 
Begin the sentence with \texttt{"A"} or \texttt{"An"}, 
avoid using any verbs (e.g., do not use “is”, “has”, etc.), 
and keep the description no longer than 10 words.
\end{promptbox}

\begin{promptbox}{Location Description Prompt}
\label{prompt:location_prompt}

Analyze the following image and provide a concise description for each object listed in the object annotations.
Object Annotations (use as reference only, focus on visible objects):
\texttt{\{obj\_ann\_str\}}.

Guidelines:

1. Treat the object annotations only as reference. If the object is not clearly visible in the image, write \texttt{"Not visible"}.  

2. Focus on the object's color, shape, size, and spatial relationships with other visible objects.  

3. Do not invent or describe objects that are not present in the image.  

4. Return your output strictly as a JSON array. Each element must follow this format:  
   \texttt{\{"ObjectID": "<id>", "Description": "<short description>"\}}  

\end{promptbox}

\begin{promptbox}{Detection Prompt}
\label{prompt:detection_prompt}

You are given an image and a list of object descriptions.  
For each object, locate it in the image and return the result as a JSON array with the following structure:
\begin{verbatim}
[
  {
    "object_id": 
        "<original_object_id>",
    "bbox_2d":
        [x_min, y_min, x_max, y_max],
    "label": 
        "<2-5 words>"
  },
  ...
]
\end{verbatim}
\end{promptbox}

\begin{promptbox}{Bbox/Seg Task Prompt}
\label{prompt:bbox_seg_prompt}

Predict the bounding box of the referred object in the image based on the instruction: \texttt{"\{\{ content | trim \}\}"}. First, output the thinking process in \texttt{<think> </think>} tags, then output the final answer in \texttt{<answer> </answer>} tags. Follow the format:
\begin{quote}
\texttt{<think> thinking process </think>}\\
\texttt{<answer>(x\_min,y\_min),
                (x\_max,y\_max)</answer>}
\end{quote}
\end{promptbox}

\begin{promptbox}{Grasp Task Prompt}
\label{prompt:grasp_prompt}

Predict a stable 2D rectangular grasp pose for the target object based on the instruction: \texttt{"\{\{ content | trim \}\}"}. First, output your reasoning process in \texttt{<think> </think>} tags, then output the final grasp pose in \texttt{<answer> </answer>} tags. Follow the format:
\begin{quote}
\texttt{<think> thinking process </think>}\\
\texttt{<answer>(x, y, theta, width)</answer>}
\end{quote}

\end{promptbox}

\begin{promptbox}{Contact Task Prompt}
\label{prompt:contact_prompt}

Predict one stable two-finger grasp contact pair (two 2D coordinates) for the target object described in the instruction: \texttt{"\{\{ content | trim \}\}"}. First, output the thinking process in \texttt{<think> </think>} tags, then output the final answer in \texttt{<answer> </answer>} tags. Follow the format:
\begin{quote}
\texttt{<think> thinking process </think>}\\
\texttt{<answer>(x1,y1),(x2,y2)</answer>}
\end{quote}
\end{promptbox}

\subsection{Details of RealVLG-R1}
\textbf{Task Prompt.} In Fig. 3 of the main text , the prompts for the Bbox and Segmentation tasks are defined in Prompt \ref{prompt:bbox_seg_prompt} , while the Grasp task and Contact task are defined in Prompt \ref{prompt:grasp_prompt} and Prompt \ref{prompt:contact_prompt}, respectively.

\textbf{Verifiable Rewards.} 
Each reward consists of two components: a \textit{format reward} $R_{\text{Format}}$ and a \textit{task reward} $R_{\text{Task}}$. 
The format reward is obtained through regular expression matching, assigning $R_{\text{Format}} = 1$ if the model output strictly follows the predefined format and $R_{\text{Format}} = 0$ otherwise. 
The task reward is computed based on task-specific evaluation metrics and normalized to the range $[0,1]$. 
The final composite reward is computed as a weighted sum:
\begin{equation}
    R(q, o) = \alpha R_{\text{Format}} + \beta R_{\text{Task}},
\end{equation}
where $\alpha = 0.1$ and $\beta = 0.9$ balance structural validity and task-level accuracy.

\section{Further Experiments}

\subsection{Qualitative Comparison of Data Quality}
\begin{figure*}[t]
    \centering
    \includegraphics[width=0.98\textwidth]{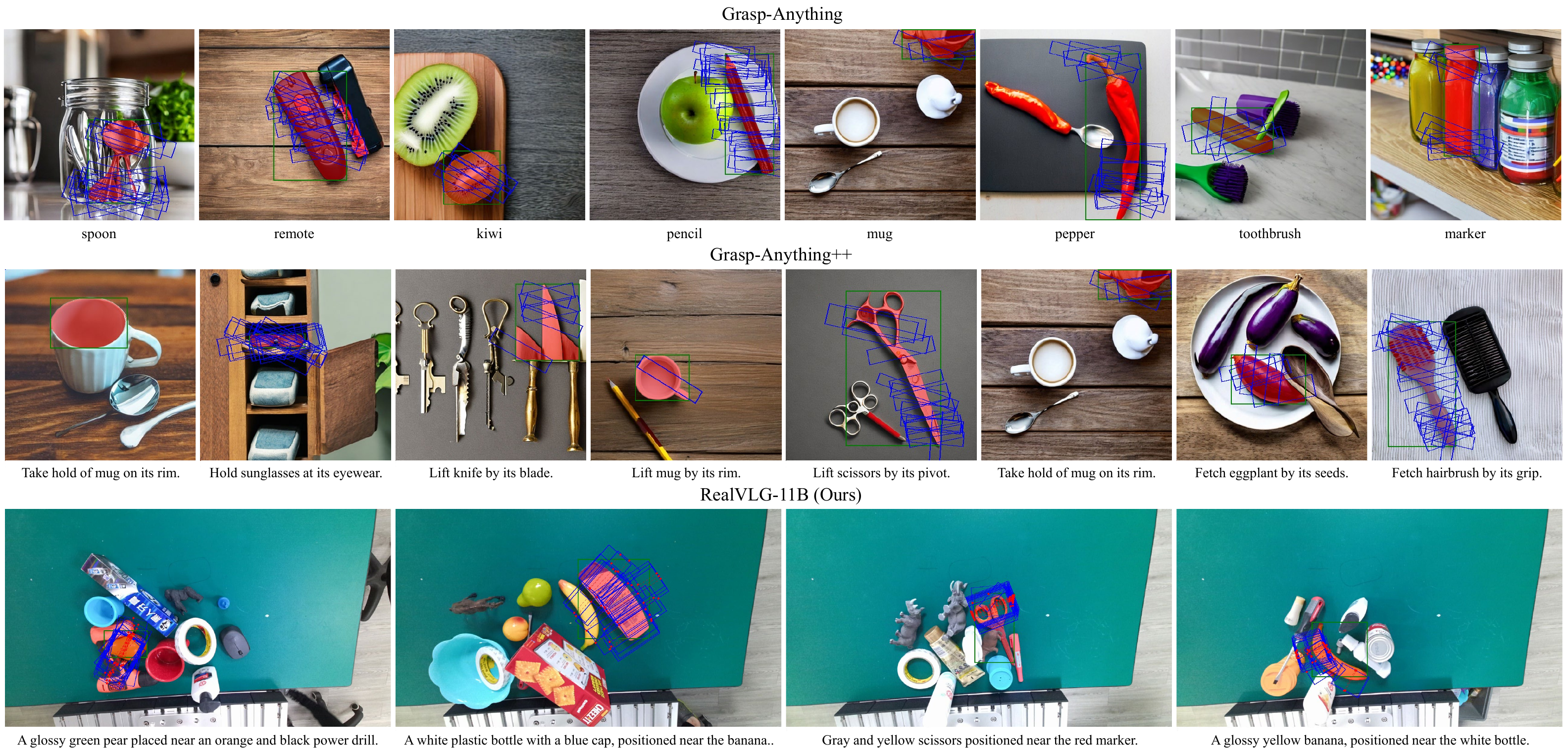}
    \caption{\textbf{Qualitative Comparison of Data Quality.} Unlike the diffusion-generated, low-resolution images and weakly aligned textual and grasp annotations in Grasp-Anything datasets, RealVLG-11B provides high-resolution real-world imagery, instance-level language grounding, and standardized, physically executable grasp labels, enabling more accurate and robust visual–language grasping.}
    \label{fig:realvlg_datacompare}
    
\end{figure*}

As illustrated in Fig. \ref{fig:realvlg_datacompare}, we compare RealVLG-11B with the Grasp-Anything family of datasets in terms of image fidelity, linguistic specificity, and the reliability of grasp annotations. The language instructions in Grasp-Anything \cite{graspanything2024} primarily describe object categories, while Grasp-Anything++ \cite{lgd2024} introduces limited part-level cues but remains restricted to highly templated and semantically simplistic expressions. Moreover, both datasets rely on diffusion-generated images with relatively low resolution (416$\times$416), which exhibit noticeable artifacts and distortions, particularly in complex geometric structures and fine-grained textures. Their textual descriptions also show weak alignment with visual content, failing to consistently correspond to specific object instances in the scene.

In terms of grasp supervision, the Grasp-Anything datasets depend on RAGT-3/3 to synthesize grasp poses, which results in high-noise, low-precision annotations that lack semantic coherence with the associated language instructions. Such limitations make it challenging to support fine-grained visual–language–action learning.

In contrast, RealVLG-11B is constructed entirely from high-resolution real-world images (1280$\times$720), preserving authentic geometric details, texture richness, and environmental variability, thereby enhancing both data realism and downstream generalization. Its linguistic annotations are produced by large vision–language models and subsequently validated by human experts, enabling instance-level grounding of objects and object parts and ensuring high-quality image–language alignment. Meanwhile, the grasp pose annotations in RealVLG-11B undergo standardized processing, yielding higher physical executability and stronger annotation accuracy. Overall, RealVLG-11B substantially surpasses the Grasp-Anything datasets in visual quality, semantic granularity, and grasp annotation reliability, providing a more robust foundation for training multimodal robotic agents capable of grounded perception and stable manipulation in real-world environments.

\subsection{Details of Baselines}

For Gemini2.5-Flash, since the model can directly output segmentation masks, the evaluation for detection and segmentation tasks follows the prompt specified in its official documentation\footnote{\url{https://ai.google.dev/gemini-api/docs/image-understanding\#python\_5}} (see Prompt \ref{prompt:gemini_prompt}).

\begin{promptbox}{Gemini2.5-Flash Evaluation Prompt}
\label{prompt:gemini_prompt}

Give the segmentation masks for the object: \texttt{"\{\{ description \}\}"} ".
Output a JSON list of segmentation masks where each entry contains the 2D
bounding box in the key ``box\_2d", the segmentation mask in key ``mask", and
the text label in the key ``label". Use descriptive labels.

\end{promptbox}

For other tasks, all baseline models use the same task prompt as RealVLG-R1.
\begin{figure}[t]
    \centering
    \includegraphics[width=0.48\textwidth]{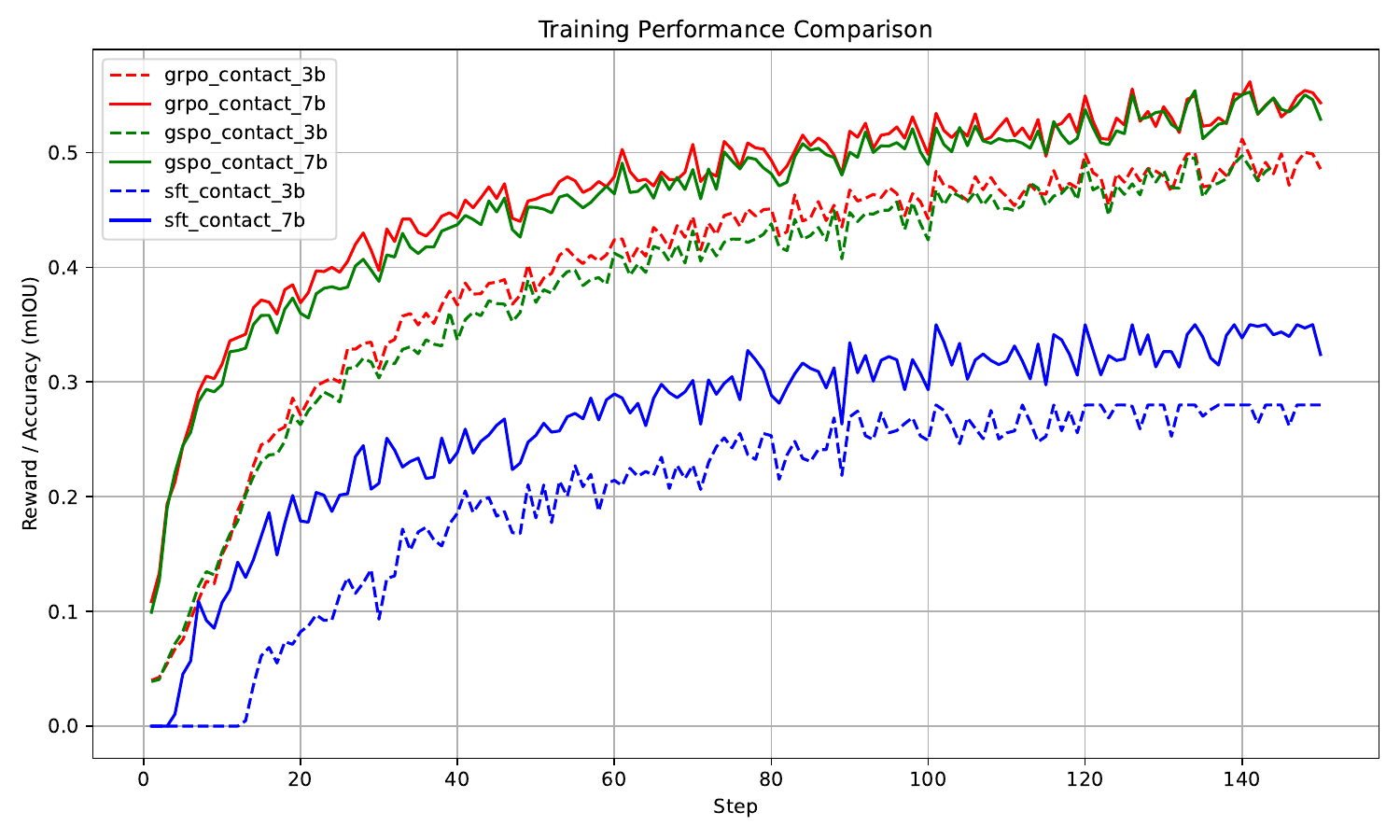}
    \caption{\textbf{Training reward/accuracy curves for GRPO, GSPO, and SFT on Contact tasks.} Overall, GRPO and GSPO significantly improve SFT through RLVR. GRPO achieves slightly higher accuracy on 3B, while GSPO performs better on 7B and exhibits more stable outputs across training steps.}
    \label{fig:realvlg-training}
    
\end{figure}

\subsection{Comparison of Optimization Strategies}
Fig. \ref{fig:realvlg-training} presents a comparative analysis of the training performance of GRPO, GSPO, and SFT on the grasp contact point prediction task, evaluated across 3B and 7B model scales. Overall, both GRPO and GSPO substantially outperform the SFT baseline by leveraging the Reward-Driven Learning with Verifiable Rewards (RLVR) paradigm, demonstrating its effectiveness in achieving precise visual grounding and grasping. Both reinforcement learning methods employ a group-wise advantage estimation strategy, which contributes to their early performance gains.

When considering model scale and optimization strategies, notable differences emerge. For the 3B models, GRPO exhibits a slight advantage due to its token-level importance weights, which provide finer-grained gradient updates. This granularity facilitates more focused optimization in parameter-limited small models, resulting in marginally higher peak accuracy. In contrast, for the 7B models, GSPO outperforms GRPO by utilizing sequence-level importance weights with length normalization. This approach mitigates gradient variance over long sequences, enabling smoother optimization and allowing larger models to fully exploit their expressive capacity.

In terms of training stability, GSPO demonstrates more consistent and stable convergence, whereas GRPO’s token-level weighting can induce higher variance in long-sequence tasks, leading to more oscillatory training behavior. These findings confirm the efficacy of RLVR in enhancing grasp contact point prediction and highlight a subtle trade-off between optimization granularity and model scale: fine-grained token-level optimization is more suitable for small models aiming for peak performance, while sequence-level optimization ensures robustness and convergence quality for large models.

\section{Evaluation on Real Robot}
\subsection{Implementation Details}
\begin{figure}[t]
    \centering
    \includegraphics[width=0.48\textwidth]{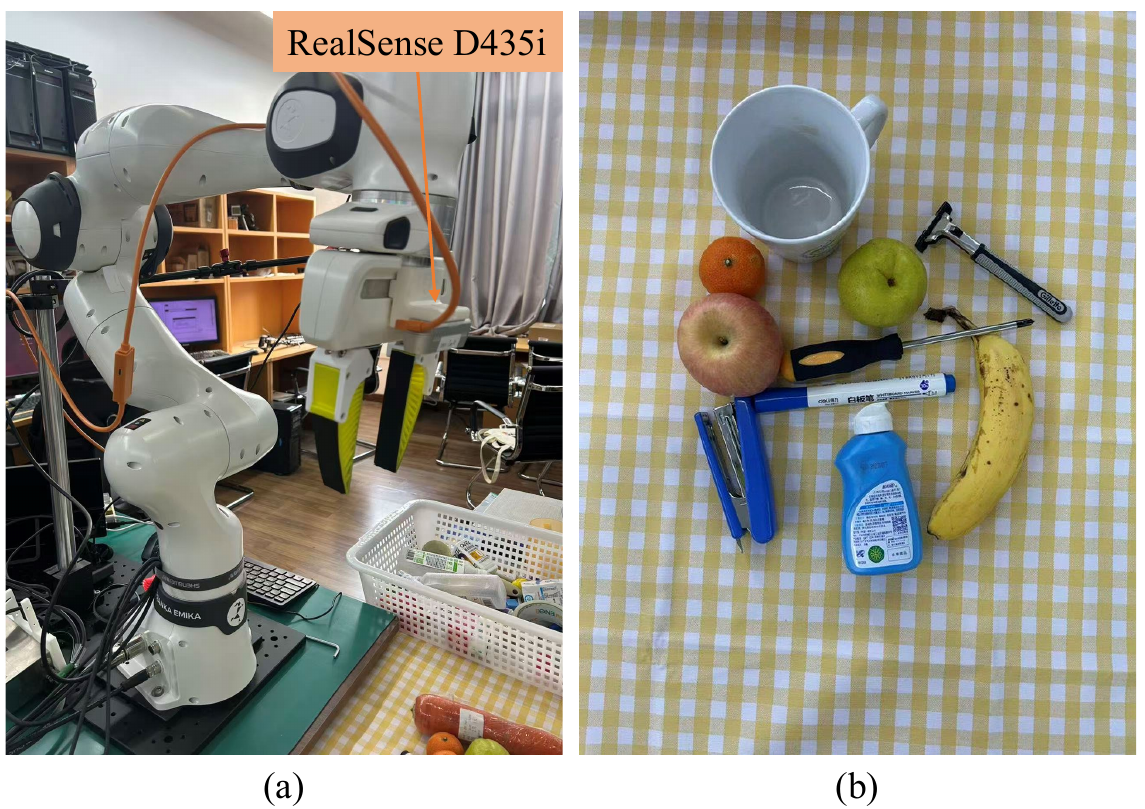}
    \caption{\textbf{Real-world experimental setup.} (a) The 7-DoF Franka Research 3 robot equipped with an eye-in-hand Intel RealSense D435i camera, used for real-world evaluation of RealVLG-R1. (b) The set of 10 test objects used to assess the model’s generalization and manipulation performance.}
    \label{fig:realvlg_setting}
\end{figure}

To assess the open-world generalization capability of our model in real-world environments, we conducted a series of robotic manipulation experiments. As illustrated in Fig. \ref{fig:realvlg_setting}, the evaluation was performed using a 7-DoF Franka Research 3 manipulator equipped with an eye-in-hand Intel RealSense D435i RGB-D camera. We designed a suite of ten grasping tasks involving diverse household objects, including \textit{Cup}, \textit{Orange}, \textit{Apple}, \textit{Pear}, \textit{Stapler}, \textit{Blue bottle}, \textit{Banana}, \textit{Marker}, \textit{Screwdriver}, and \textit{Razor}. Each task was executed 10 times, and the average success rate was reported.

\subsection{Experimental Settings}
To comprehensively evaluate the generalization capability of our proposed RealVLG-R1 model in real-world environments, we designed experiments across two complementary settings and selected two representative baselines for comparison. In the \textit{Single} setting, only the target object is placed to ensure stable operation of the classical geometric grasping baseline GraspNet \cite{graspnet2020}, which lacks language-conditioned capabilities. This setting aims to assess RealVLG-R1’s performance in basic geometric grasping accuracy. In the more challenging \textit{Clutter} setting, all 10 objects are placed simultaneously, and the robot is required to sequentially grasp the specified targets according to language instructions. In this scenario, RealVLG-R1 is compared only with LGD \cite{lgd2024}, a baseline that supports language-conditioned grasping.  \textbf{The comparison is conducted using the RealVLG-R1 3B model trained with GRPO, which predicts grasp contact points.} This hierarchical evaluation strategy allows us to clearly demonstrate RealVLG-R1’s semantic reasoning advantages over purely geometric baselines, while highlighting its zero-shot deployment performance and cross-modal generalization capability in highly complex cluttered scenes.

\subsection{Experimental Results}
\begin{table*}[ht]
    \centering
    \resizebox{0.96\textwidth}{!}{
    \begin{tabular}{l|cccccccccc|c}
    \hline Method & Cup & Orange & Pear & Apple  &  Banana & Stapler & BlueBottle & Marker & Screwdriver & Razor & Average \\
    \hline
    GraspNet \cite{graspnet2020} & 10\% & 60\% & 70\% & 60\% & 30\% & 50\% & 40\% & 20\% & 30\% & 10\% & 38\% \\
    
    RealVLG-R1 (Ours) & \textbf{90\%} & \textbf{100\%} & \textbf{100\%} & \textbf{100\%} & \textbf{100\%} & \textbf{80\%} & \textbf{70\%} & \textbf{80\%} & \textbf{100\%} & \textbf{90\%} & \textbf{81\%} \\
    \hline
    \end{tabular}
}
    \caption{\textbf{Quantitative real-world grasping results in the Single setting.} RealVLG-R1 performs language-conditioned grasping, whereas GraspNet serves as a vision-only baseline. Each task is executed 10 times.}
    \label{tab:tab_single}
\end{table*}

\begin{figure*}[t]
    \centering
    \includegraphics[width=0.98\textwidth]{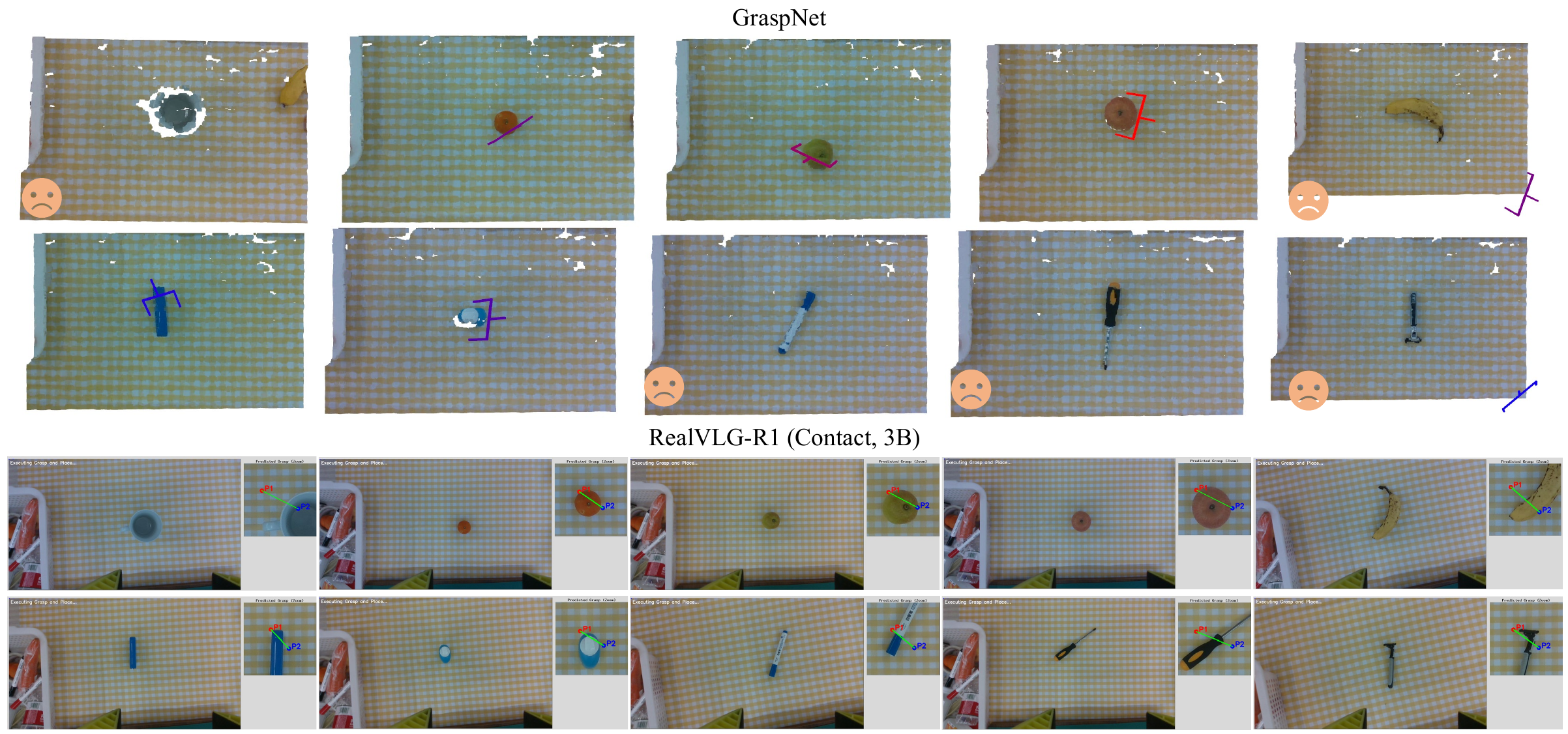}
    \caption{\textbf{Qualitative real-world grasping results in the Single setting.} GraspNet often fails or predicts misaligned grasp poses due to noisy or incomplete point cloud data (e.g., \textit{Cup}), reflective surfaces, and small or thin objects, such as \textit{Marker}, \textit{Screwdriver}, and \textit{Razor}. In contrast, RealVLG-R1 leverages RGB vision and language instructions to accurately localize the target and generate executable grasp contact points, demonstrating robust and reliable grasping behavior across diverse objects.}
    \label{fig:realvlg_single}
    
\end{figure*}

\textbf{Grasping Results in the Single Setting.}
As shown in Table \ref{tab:tab_single} and Fig. \ref{fig:realvlg_single}, RealVLG-R1 exhibits a substantial performance advantage over GraspNet in the Single setting. Although GraspNet is capable of predicting 6-DoF grasp poses, its effectiveness in real-world scenarios is heavily constrained by the quality of the reconstructed point cloud. When objects contain reflective surfaces, sparse textures, or complex geometries, the depth measurements often become incomplete or noisy. For instance, for objects such as the \textit{Cup}, the material and surface properties frequently lead to partial or degraded point cloud reconstructions, preventing GraspNet from producing any feasible grasp pose. Furthermore, for small and slender objects (such as the \textit{Marker}, \textit{Screwdriver}, and \textit{Razor}), their point clouds tend to merge with the tabletop during reconstruction, making them nearly indistinguishable in the depth domain. As a result, GraspNet commonly predicts grasp poses that fall outside the true object region, ultimately causing grasp failures.

In contrast, RealVLG-R1 relies solely on RGB visual information and leverages language-conditioned grounding to accurately localize the target object. The model predicts grasp contact points directly in the image domain, and as long as the depth along the line connecting these contact points is valid, a physically executable grasp can be produced. This design effectively circumvents the depth degradation issues that commonly arise in real-world sensing (such as missing geometry, noise, or surface ambiguities), allowing RealVLG-R1 to maintain robust grasp performance across diverse visual conditions.

Overall, RealVLG-R1 achieves a significantly higher mean success rate than GraspNet in real-world single-object grasping, yielding a 43\% improvement, which underscores the robustness and practical effectiveness of a vision-based, language-driven grasping framework in real-world environments.

\begin{table*}[ht]
    \centering
    \resizebox{0.96\textwidth}{!}{
    \begin{tabular}{l|cccccccccc|c}
    \hline Method & Cup & Orange & Pear & Apple  &  Banana & Stapler & BlueBottle & Marker & Screwdriver & Razor & Average \\
    \hline
    LGD \cite{lgd2024} & 0\% & 0\% & 0\% & 0\% & 0\% & 20\% & 0\% & 0\% & 0\% & 0\% & 2\% \\
    
    RealVLG-R1 (Ours) & \textbf{70\%} & \textbf{90\%} & \textbf{80\%} & \textbf{90\%} & \textbf{70\%} & \textbf{90\%} & \textbf{70\%} & \textbf{80\%} & \textbf{70\%} & \textbf{80\%} & \textbf{79\%} \\
    \hline
    \end{tabular}
}
  \caption{\textbf{Quantitative real-world grasping results in the Clutter setting.} The table reports the grasp success rates of RealVLG-R1 and the baseline LGD across 10 target objects. Results highlight RealVLG-R1's superior zero-shot performance in cluttered multi-object environments, whereas LGD fails to reliably perform language-conditioned grasping.}
    \label{tab:tab_clutter}
\end{table*}

\begin{figure*}[t]
    \centering
    \includegraphics[width=0.98\textwidth]{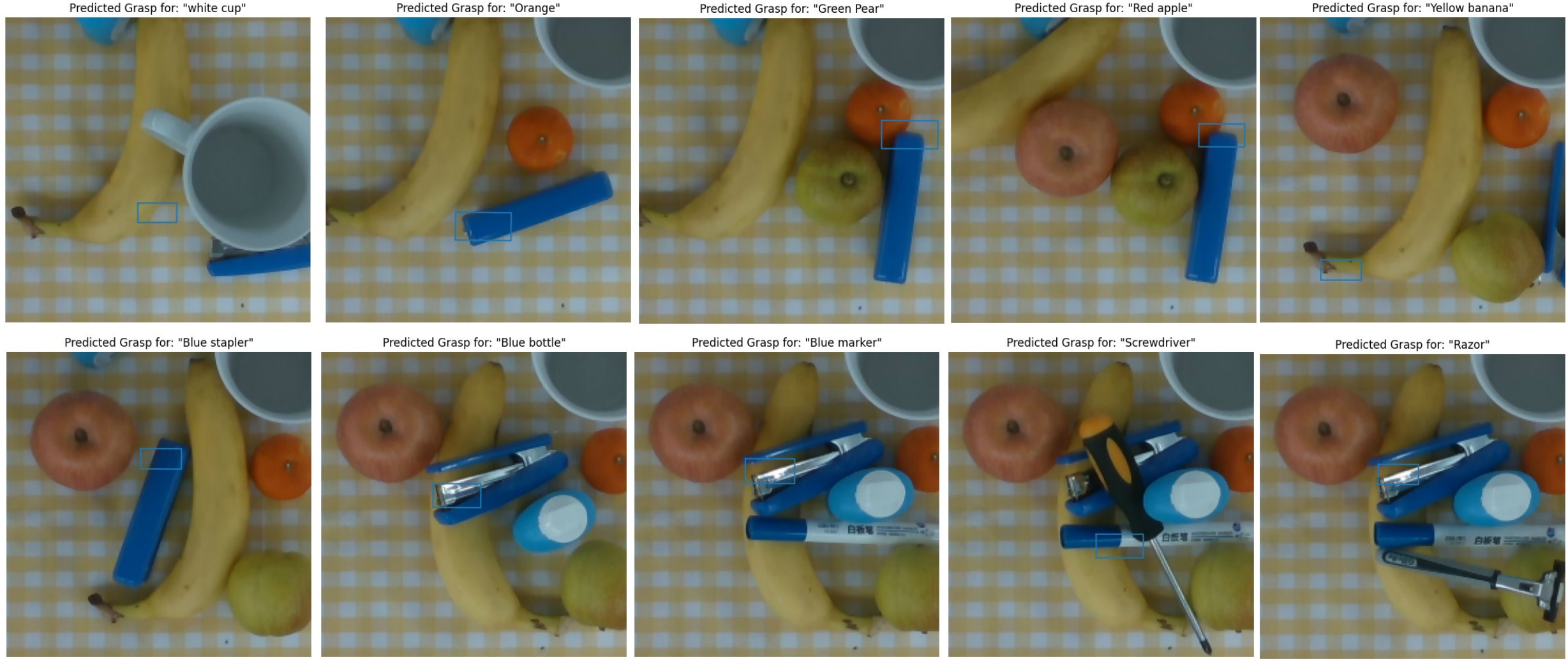}
    \caption{\textbf{Qualitative real-world grasping results of LGD \cite{lgd2024} in the Clutter setting.} LGD struggles to perform language-conditioned grasps in cluttered environments due to limited perceptual resolution, suboptimal language integration, and reliance on unconditional grasp pose predictions.}
    \label{fig:realvlg_clutter_lgd}
    
\end{figure*}

\textbf{Grasping Results in the Clutter Setting.}
As shown in Table \ref{tab:tab_clutter}, in the highly challenging Clutter Setting, our proposed RealVLG-R1 demonstrates overwhelming performance advantages. The model achieves a grasp success rate of 70\% or higher across all 10 target objects, with an average success rate of 79\%. These results strongly indicate that RealVLG-R1 is capable of accurately identifying, localizing, and executing grasping tasks according to natural language instructions, even in environments characterized by high object density and cluttered interference. Although certain grasp contact points may occasionally exhibit instability, suggesting that further enhancement is needed for real-world deployment, the overall performance clearly demonstrates the model’s strong zero-shot generalization capability.

In contrast, the baseline LGD achieves an average success rate of only 2\% in real-world cluttered scenes, highlighting its failure in language-conditioned grasping tasks. This significant gap arises from inherent design limitations and training data constraints: the network, based on the GGCNN \cite{morrison2018closing} architecture, processes images of only $224 \times 224$ pixels, resulting in a severely restricted perceptual field that hampers accurate recognition of object context and complete spatial information in densely cluttered environments. Although LGD attempts to incorporate language instructions through CLIP \cite{radford2021learning} embeddings, experimental results indicate that such integration is suboptimal. As illustrated in Fig. \ref{fig:realvlg_clutter_lgd}, LGD rarely predicts grasp poses corresponding to specific objects based on language instructions, with its output grasp poses largely independent of the provided commands. Furthermore, the ``accuracy” reported for LGD on Grasp-Anything++ dataset predominantly reflects the unconditional grasp pose prediction capabilities inherited from GGCNN, rather than poses specified by language instructions. In the absence of high-quality language-to-pose aligned training data, LGD’s language-driven object grasping success in real-world scenes is effectively negligible, with the few successful cases (e.g., a 20\% success rate for the \textit{Stapler}) attributable to geometric cues alone.

\begin{figure*}[t]
    \centering
    \includegraphics[width=0.75\textwidth]{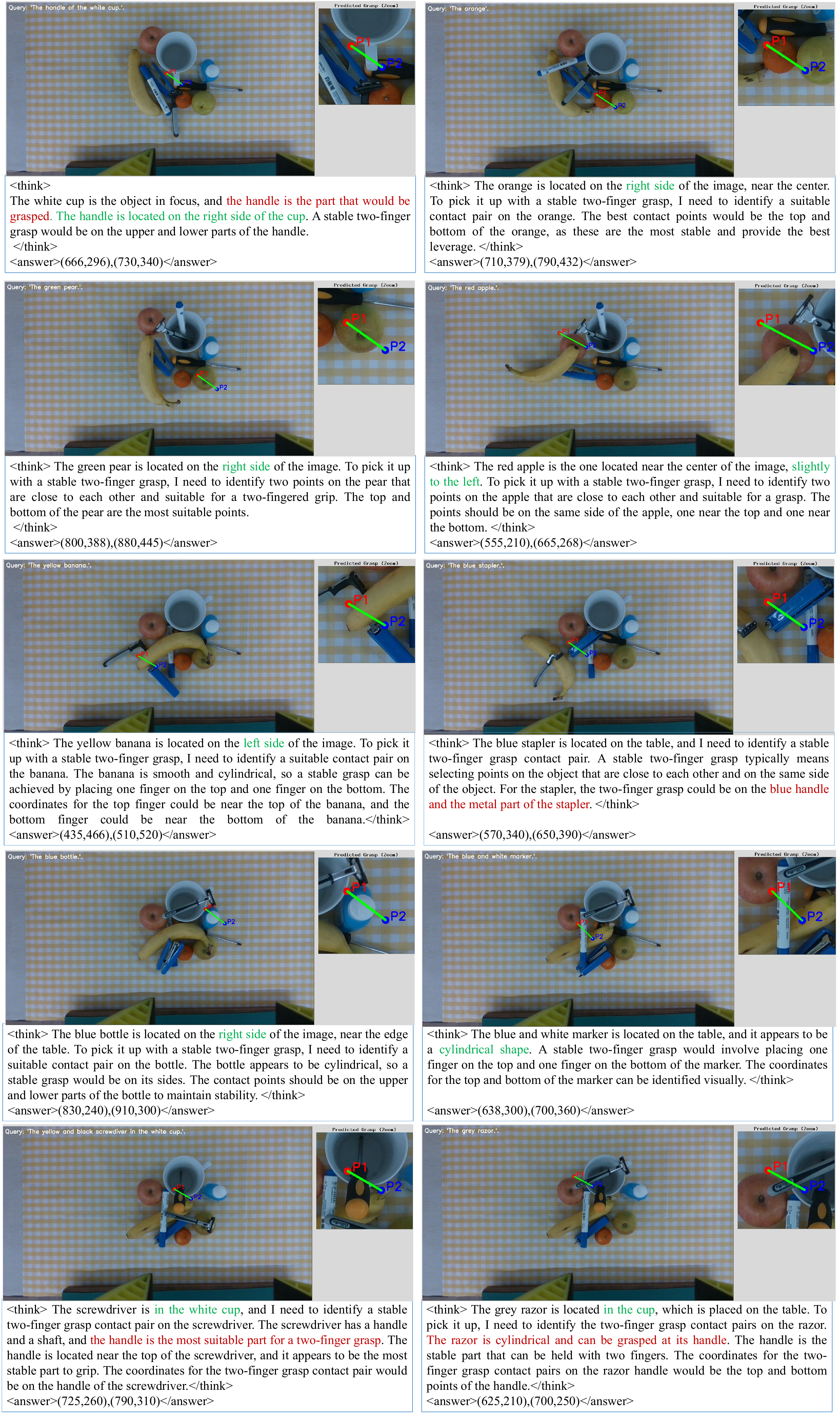}
    \caption{\textbf{Qualitative real-world grasping results of RealVLG-R1 in the Clutter setting.} RealVLG-R1 demonstrates accurate language-conditioned grasping, robust zero-shot performance in cluttered environments, and interpretable predictions of grasp poses.}
    \label{fig:realvlg_clutter_ours}
\end{figure*}

In sharp contrast to LGD’s limitations, RealVLG-R1 exhibits precise and reliable language-conditioned grasping capability, enabling the model to accurately localize target objects in wide-field, multi-object scenarios. Moreover, as shown in Fig. \ref{fig:realvlg_clutter_ours}, RealVLG-R1 provides strong reasoning capabilities, offering compelling interpretability for predicted grasp poses. This reasoning mechanism not only enhances model transparency and reliability but also provides an additional internal validation of grasp pose accuracy. By combining precise perception, accurate language alignment, and interpretable reasoning, RealVLG-R1 achieves high grasping success rates in cluttered environments, fully demonstrating its zero-shot deployment performance and cross-modal generalization capabilities in complex real-world scenarios.

\end{document}